\documentclass[acmsmall]{acmart}
\usepackage{dutchcal}
\usepackage{tabularx}
\usepackage{multirow}
\usepackage{makecell}
\usepackage{pifont}
\usepackage{threeparttable}
\usepackage{color}
\AtBeginDocument{%
  \providecommand\BibTeX{{%
    \normalfont B\kern-0.5em{\scshape i\kern-0.25em b}\kern-0.8em\TeX}}}

\acmJournal{JACM}
\acmVolume{37}
\acmNumber{4}
\acmArticle{111}
\acmMonth{8}

\begin{document}

%%
%% The "title" command has an optional parameter,
%% allowing the author to define a "short title" to be used in page headers.
\title{Inference Attacks: A Taxonomy, Survey, and Promising Directions}

%%
%% The "author" command and its associated commands are used to define
%% the authors and their affiliations.
%% Of note is the shared affiliation of the first two authors, and the
%% "authornote" and "authornotemark" commands
%% used to denote shared contribution to the research.
\author{Feng Wu}
\affiliation{%
	\institution{University of Technology Sydney}
	\city{Sydney}
	\country{Australia}}
\email{gzwf4091@163.com}

\author{Lei Cui}
\affiliation{%
	\institution{Shandong Fundamental Research Center for Computer Science}
	\city{Jinan}
	\country{China}}
\email{Alencui@outlook.com}

\author{Saowen Yao}
\affiliation{%
	\institution{Yunnan University}
	\city{Kunming}
	\state{Yunnan}
	\country{China}}
\email{ yaosw@mail.ynu.edu.cn}

\author{Shui Yu}
\affiliation{%
	\institution{University of Technology Sydney}
	\city{Sydney}
	\country{Australia}}
\email{shui.yu@uts.edu.au}

%%
%% By default, the full list of authors will be used in the page
%% headers. Often, this list is too long, and will overlap
%% other information printed in the page headers. This command allows
%% the author to define a more concise list
%% of authors' names for this purpose.
%\renewcommand{\shortauthors}{Trovato and Tobin, et al.}

%%
%% The abstract is a short summary of the work to be presented in the
%% article.
\begin{abstract}

The prosperity of machine learning has also brought people's concerns about data privacy. Among them, inference attacks can implement privacy breaches in various MLaaS scenarios and model training/prediction phases. Specifically, inference attacks can perform privacy inference on undisclosed target training sets based on outputs of the target model, including but not limited to statistics, membership, semantics, data representation, etc. For instance, infer whether the target data has the characteristics of AIDS. In addition, the rapid development of the machine learning community in recent years, especially the surge of model types and application scenarios, has further stimulated the inference attacks' research. Thus, studying inference attacks and analyzing them in depth is urgent and significant. However, there is still a gap in the systematic discussion of inference attacks from taxonomy, global perspective, attack, and defense perspectives. This survey provides an in-depth and comprehensive inference of attacks and corresponding countermeasures in ML-as-a-service based on taxonomy and the latest researches. Without compromising researchers' intuition, we first propose the 3MP taxonomy based on the community research status, trying to normalize the confusing naming system of inference attacks. Also, we analyze the pros and cons of each type of inference attack, their workflow, countermeasure, and how they interact with other attacks. In the end, we point out several promising directions for researchers from a more comprehensive and novel perspective.

\end{abstract}

%%
%% The code below is generated by the tool at http://dl.acm.org/ccs.cfm.
%% Please copy and paste the code instead of the example below.
%%
\begin{CCSXML}
<ccs2012>
 <concept>
  <concept_id>10010520.10010553.10010562</concept_id>
  <concept_desc>General and reference~Surveys and overviews</concept_desc>
  <concept_significance>500</concept_significance>
 </concept>
 <concept>
  <concept_id>10010520.10010575.10010755</concept_id>
  <concept_desc>Computing methodologies~Machine learning</concept_desc>
  <concept_significance>300</concept_significance>
 </concept>
 <concept>
  <concept_id>10010520.10010553.10010554</concept_id>
  <concept_desc>Security and privacy~Privacy-preserving protocols</concept_desc>
  <concept_significance>100</concept_significance>
 </concept>
</ccs2012>
\end{CCSXML}

\ccsdesc[500]{General and reference~Surveys and overviews}
\ccsdesc[300]{Computing methodologies~Machine learning}
\ccsdesc[100]{Security and privacy~Privacy-preserving protocols}

%%
%% Keywords. The author(s) should pick words that accurately describe
%% the work being presented. Separate the keywords with commas.
\keywords{MLaaS, privacy-preserving, inference attacks, inference attack countermeasures}

%%
%% This command processes the author and affiliation and title
%% information and builds the first part of the formatted document.
\maketitle

\section{Introduction}
Machine learning has shown data processing capabilities far beyond humans' in many fields, such as medical diagnosis, financial risk control, and industrial production \cite{jordan2015machine}. However, building a machine learning model requires innumerous computing resources and data, which is unaffordable for most enterprises and individuals. Therefore, some companies that can build machine learning models provide their models' interface as cloud services to those who cannot build models,  forming Machine Learning as a Service (MLaaS). MLaaS was first conceptualized by Ribeiro et al. in $2015$ \cite{ribeiro2015mlaas}, and its scalable, flexible, and non-blocking platform features enable it to have large-scale applications in the industry \cite{hesamifard2018privacy,li2017scaling}. With the advancement of machine learning technology and ethical thinking, privacy-preserving have become more prominent \cite{tseng2020compressive}. Thus, recently, MLaaS architectures with privacy-preserving (such as federated learning) have emerged. These architectures solve the privacy problems caused by the traditional MLaaS to a certain extent, e.g., cloud servers no longer collecting users' private data for model training\cite{wu2021data}.

However, many studies have shown that machine learning models are prone to remembering information of training data in training phase \cite{carlini2019secret,song2017machine,zhang2021understanding}. Many privacy attacks against MLaaS have subsequently been discovered based on the fact, and one of the most threatening attack type is inference attack. Conventionally, inference attacks can be divided into membership inference attacks (MIAs) \cite{shokri2017membership}, property inference attacks (PIAs) \cite{ganju2018property}, model inversion attacks (MIs) \cite{fredrikson2015model}, and model inference attacks \cite{tramer2016stealing}. \textbf{Membership inference attacks (MIAs)} aim to determine whether a record exists in the target model's training set (i.e., cloud model). For example, attackers can use membership inference attacks to infer the customer intersection of they and their business competitors to formulate targeted business plans. \textbf{Property inference attacks (PIAs)} mainly leverage publicly available information of a specific instance to infer the unknown information of the instance (such as sexual orientation, political tendency, gender, etc.) in online social networks. Subsequently, the concept of PIA was extended to infer the global properties of the training set (such as gender ratio, age ratio, etc.) and gradually became mainstream. For \textbf{model inversion attacks (MIs)}, it focuses on reconstructing data in the training set of machine learning models, such as a training image, a sentence, etc. For example, in federated learning, a malicious central server can use model inversion attacks to steal data from clients. Unlike the above-mentioned inference attacks aim at training data's privacy, \textbf{model inference attacks (also known as model extraction model, MEAs)} aim to obtain a duplicate of the cloud model through outputs of the cloud model. Since building a cloud model is a costly business behaviour, and the cloud model is an important and private business asset, the threat of model inference attacks cannot be ignored. 

Privacy security have a non-negligible impact on machine learning related industrial deployment, thus, many surveys have summarized plenty data/model privacy issues for MLaaS.  \cite{shokri2017membership,nasr2019comprehensive,wang2021membership,olatunji2021membership,duddu2020quantifying} are instances of membership inference attacks in scenarios such as federated learning, social networks, and graph neural networks. \cite{kesarwani2018model,bastani2017interpreting,jia2021entangled} emphasized that cloud services are incredibly vulnerable to model inference attacks. The threat posed by property inference attacks was demonstrated by \cite{liu2012protecting,zhang2022inference,gong2016you} through online social networks. Compared with other threats (such as adversarial attacks \cite{goodfellow2014explaining}), inference attacks are one of the main threats based on deep learning technology just discovered in recent years, thus require more in-depth research. Existing surveys on inference attacks mainly summarize a specific type of inference attack in a specific scenario (such as federated learning, social network, graph neural network, etc.)\cite{hu2021membership,nasr2019comprehensive,parisot2021property}, or no focus on taxonomy in inference attacks, which makes it difficult for researchers to research from a clearer perspective. 

Unlike most surveys limited to one scenario or one type of attack, we have selected more than 200 related literature of the entire inference attack family for analysis, aiming to provide readers with a global perspective to facilitate their research. Besides, since the researches on inference attacks are still in infancy, some literature has different names for a same type of attack (e.g., Attribute inference attacks are also called model inversion attacks, but model inversion attacks have totally different meaning in computer vision \cite{fredrikson2015model,khosravy2021model,hu2021membership}; Property inference attacks and attribute inference attacks represent different concepts but have similar names\cite{hu2021membership,jia2018attriguard}, etc.). These irrationalities may confuse researchers. Therefore, based on the definition of the word "inference" and the development of inference attacks, this survey proposes a taxonomy of inference attacks called $3$MP that is more in line with researchers' intuition. According to $3$MP, we present each inference attack's implementation, research status, pros/cons, and corresponding countermeasures. Then, we combine existing privacy-preserving research and future developments to provide researchers with several valuable research directions. The main contributions of this paper are:

\textbf{1. Clearer and More Reasonable Taxonomy: } We point out the irrationality of existing taxonomies in current inference attacks researches, and propose a novel taxonomy named $3$MP to reconstruct the existing inference attacks' family structure, so that various concepts of inference attacks are more in line with research intuition, and various attack definitions are clearer. Theoretically, this taxonomy is beneficial for researchers understand and distinguish the differences between different inference attacks.

\textbf{2. Comprehensive Perspective: }To the best of our knowledge, this is the first work to discuss all types of inference attacks and provide a global view based on taxonomy. For each type of attack, we present their principles, advantages and disadvantages, countermeasures based on up-to-date literature, aiming aims to provide reference and inspiration for researchers.

\textbf{3. Future Challenges and Promising Directions: }The study of inference attacks is crucial for machine learning's application, and it is a long-term hot topic. Therefore, we summarize the unsolved challenges and promising research directions in inference attack and inference defense, respectively, based on the existing works, to inspire the follow-up research in this domain.

The rest of the paper is organized as follows. Section $2$ introduces MLaaS preliminaries. In Section $3$, we introduce the proposed taxonomy $3$MP. In Section $4$, we discuss various inference attacks and why they can work on machine learning model based on $3$MP. Section $5$ provides the corresponding countermeasures and their operation mechanism. We discuss the challenges and propose the future directions in Section $6$ and the conclusion is in Section $7$.

\section{Preliminaries about MLaaS}
\begin{figure}[h]
	\centering
	\includegraphics[width=\linewidth]{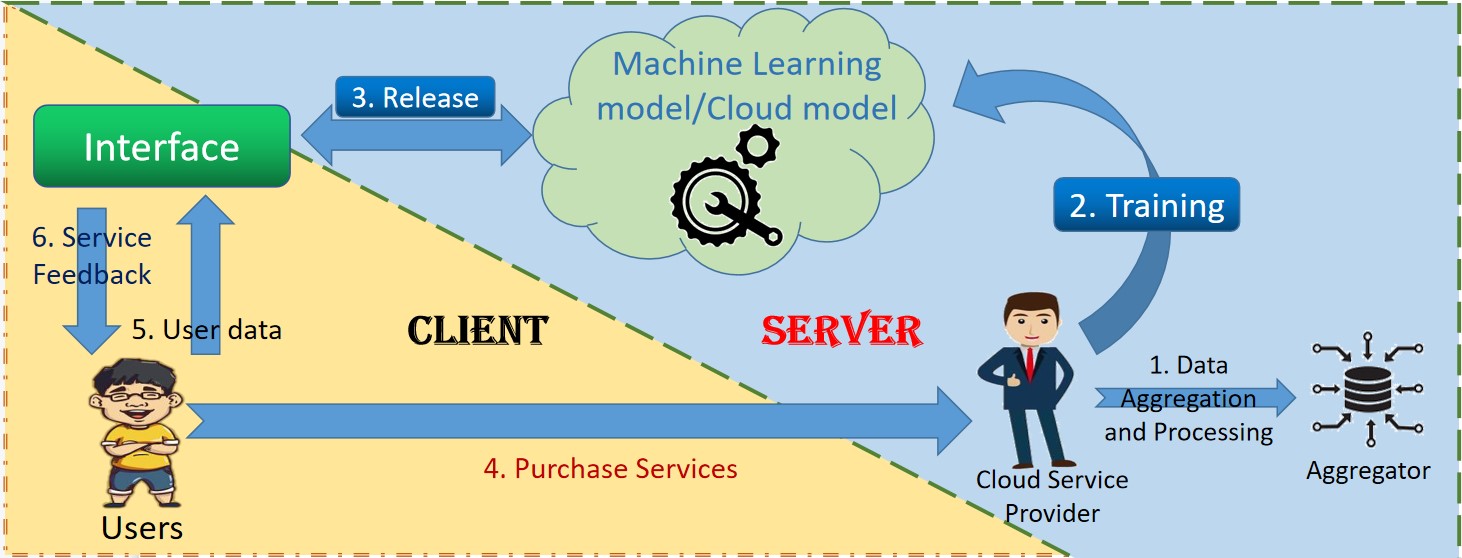}
	\caption{A overview of MLaaS: Users can enjoy machine learning services without paying attention to any specific implementation of the machine learning model.}
	\label{fig1}
\end{figure}

This section provides an overview of Machine Learning as a Service (MLaaS) and its important derivative federated learning to help readers understand how inference attacks work in MLaaS. Note that this is not a comprehensive present of MLaaS. Readers can refer to \cite{ribeiro2015mlaas,kim2018nsml,hunt2018chiron,hesamifard2018privacy,philipp2020machine} for a systematic study.
MLaaS was first mentioned in
\cite{ribeiro2015mlaas} by Ribeiro \emph{et al.}. Their purpose is to design a machine learning service platform that can be quickly used by enterprises and individuals who lack machine learning resources (such as data, computing power, etc.). The original MLaaS focused on modelling predictive models and its scalable and flexible make it can 
support different data sources. Under this architecture, the training data is received and preprocessed by a data aggregator, and the modelling component is used to train the machine learning model. In the end, service providers provide users with an interface to obtain machine learning services. Figure \ref{fig1} shows a general process of MLaaS.

MLaaS already has many applications in real life, such as intelligent apps, shopping recommendations, auxiliary medical treatment, etc. Nevertheless, MLaaS has two evident and noteworthy privacy loopholes in terms of data flow from the perspective of privacy: $(1)$. In the data aggregation phase, cloud service providers need to collect data from individuals, which brings privacy risks to individuals. $(2)$. In the service release stage, users need to upload their local data to the cloud server to obtain services (such as prediction, analysis, etc.), which also brings privacy risks to users.

\begin{figure}[h]
	\centering
	\includegraphics[width=0.9\linewidth]{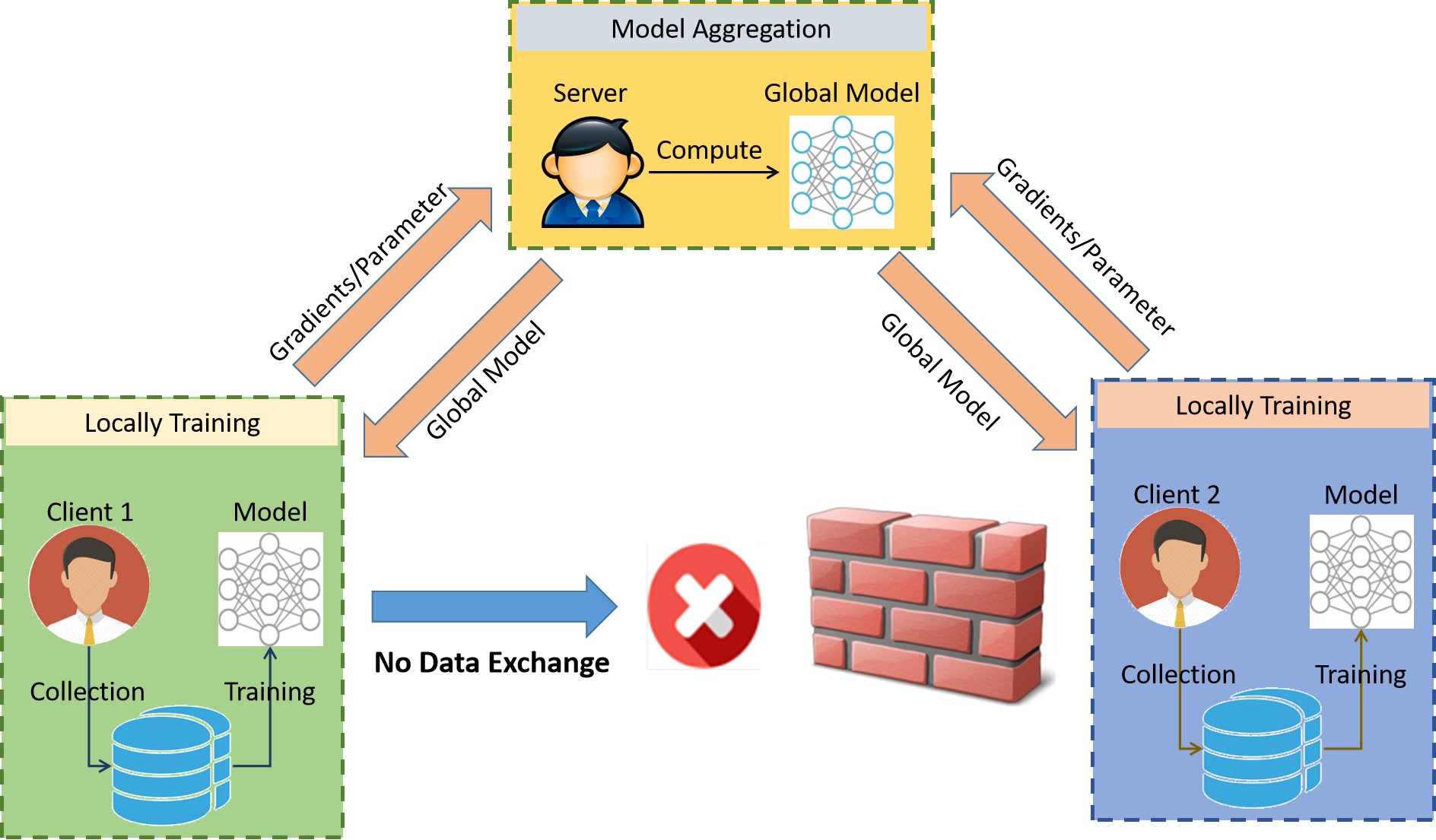}
	\caption{Architecture for a federated learning system.}
	\label{fig2}
\end{figure}

As governments and individuals pay more and more attention to privacy security, some new MLaaS service architectures, such as federated learning, have emerged in response to the above vulnerabilities. Federated learning (FL) replaces the original data aggregation process with the upload of gradients/parameters, which protects users' privacy and trains effective machine learning models. The diagram of federated learning is shown in Figure \ref{fig2}.

%我们以具有两个客户端的联邦学习系统为例。假设客户端1和客户端2持有的数据集分别为X1和x2,一般来说，x1X1=0.在一个典型的联邦学习系统中，客户端的数据量往往不足以支撑一个有效的机器学习模型的训练。为了隐私考虑，客户端也不想将其本地数据直接交由其他客户端和云端用于训练模型。因此，在与其他客户端对模型结构达成共识后，各个客户端只需要使用本地数据训练一个本地模型，再将该模型的梯度或参数上传到云端进行聚合得到一个全局模型，如式1(这是一个初级的聚合方法，如果读者有兴趣了解其他高阶的聚合方法，请参考XXXX.)。
%
%
%云服务器会将聚合后的模型分发给各个客户端进行再训练，不断重复该过程直至训练收敛。FL很好的解决了一部分传统MLaaS中的隐私问题，但也带来了新的隐私问题，我们将再第4章具体介绍。
To facilitate readers' understanding, we take a FL system with two clients as an example. Assume that the dataset held by client $1$ and client $2$ are $X_1=\{x^1_1,x^1_2,\dots,x^1_n\}$ and $X_2=\{x^2_1,x^2_2,\dots,x^2_n\}$, respectively. Generally speaking, $X_{1} \cap X_{2}=\emptyset$. Single clients' data is often insufficient to support a practical machine learning model's training in a federated learning system. For privacy reasons, clients also do not want to hand over their local data directly to other clients or the cloud server for training models. Therefore, after reaching a consensus on the model structure with other clients, each client only needs to utilise their local data to train a local model. Then upload the gradients or parameters of the local model to the server for aggregation to obtain a practical global model. The aggregation method is as shown in Equation \ref{eq1}, which is an elementary aggregation method. If readers are interested in learning advanced aggregation methods, such as aggregation with privacy-preserving or incentives, please refer to \cite{yang2020federated,blum2021one,kairouz2021distributed,jia2021blockchain,kong2021privacy}.
\begin{equation}
	\nabla W_{g l o b a l}=\frac{1}{n} \sum_{i=1}^{n} \nabla W_{i}.
	\label{eq1}
\end{equation}
After the aggregation is finished, the cloud server delivers the aggregated model (global model) to each client for retraining, and repeats this process until the training converges. FL solves some privacy issues in traditional MLaaS very well, but it also brings some new privacy issues, which we will introduce in Chapter $4$.

\section{proposed taxonomy 3MP}
To help readers better understand this survey, in this section, we first briefly introduce the current definitions of each inference attack through clear attack examples, then propose a new taxonomy 3MP based on the deficiencies of the existing taxonomy.
\subsection{Brief Introduction to inference attakcs}
\label{sec3.1}
\begin{figure}[h]
	\centering
	\includegraphics[width=0.4\linewidth]{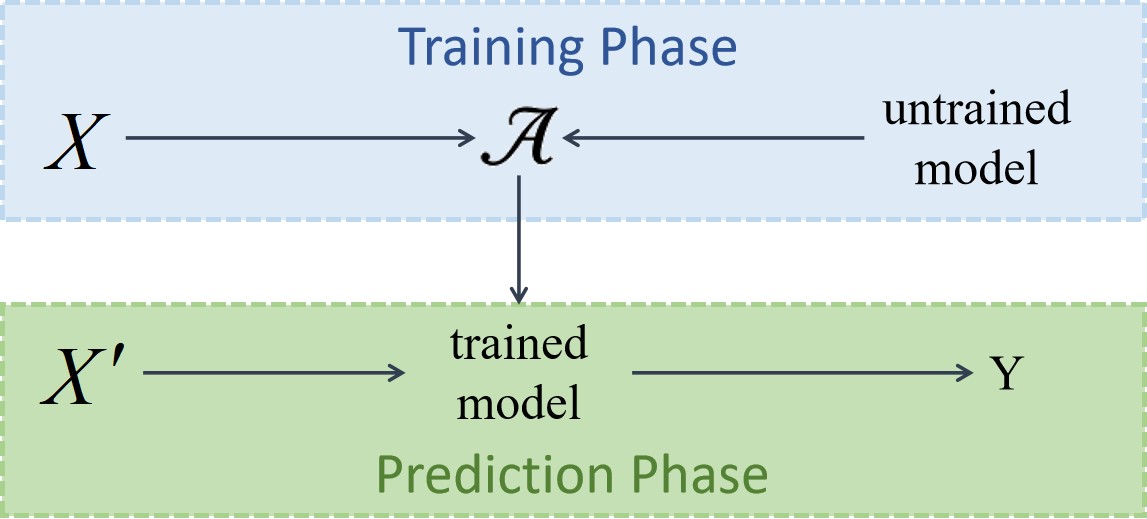}
	\caption{Different phases of model life circle.}
	\label{fig3}
\end{figure}

Currently, known inference attacks can occur during both training and inference (prediction) phase. As shown in Figure \ref{fig3}, where $\mathbcal{A}$ represents a model training algorithm, $X$ is the training set, $X^\prime$ represents the unseen data, and $Y$ is the prediction result. Next, we briefly introduce the basic definition of each attack based on the different phases as the basis for the taxonomy $3$MP.

\noindent\emph{\textbf{3.1.1 Attack in training and prediction phase}}
\begin{itemize}
\item Membership Inference Attacks (MIAs):

\textbf{Definition:} Given a specific input $x^\prime$, the attacker infers whether $x^\prime$ exists in training set $X$ through target model.

\textbf{Example:} In the era of big data, data information (such as customer information) owned by competing companies often overlaps. For example, people who use Amazon to buy products may also buy products on eBay. Generally speaking, these enterprises do not know whether their customers exist in competitors' customer sets. Moreover, they will not share their data with each other, which provides a basis for healthy business competition. However, MIAs can allow a malicious enterprise to the intersection of its customer set and its competitors through the machine learning model that provided by competitors for customers, so as to make targeted business decisions and disrupt the market order.

\item Property Inference Attacks (PIAs):

\textbf{Definition:} Given a target model trained from dataset $X$ and a pair of properties $P$ and $\bar{P}$ that an attacker is interested in (e.g., in a human face dataset, $P$ represents a male-to-female ratio of $2:1$, then $\bar{P}$ can represent a male-to-female ratio of $1:1$). Generally speaking, $P$ and $\bar{P}$ do not represent individual data information but the training set's global properties \cite{ateniese2013hacking}. An attacker can use PIAs to determine which of $P$ and $\bar{P}$ matches the global property of the training set.

\textbf{Example:} Suppose a malware classifier model is trained using the execution traces of malware and benign software. An attacker might use PIAs to obtain properties of the target model testing environment, enabling avoid detection or identify vulnerabilities.
Because the test environment affects the traces of all software, it can be considered a property of the entire training set rather than an individual record. As another example, recent research has found that underrepresentation of certain classes of people, such as women and people of color, in different datasets can lead to different performance across classes for the same classifier.

\item Attribute Inference Attacks (AIAs):

\textbf{Definition:} Given user data $A=\{a_1,a_2,a_3\}$, where $a_i$ represents the values of different attributes in the user data. Assuming that $a_1$ and $a_2$ are public, anyone can access them, and $a_3$ is private information that users do not want to disclose. The purpose of AIAs is to infer the privacy attribute $a_3$ through the information disclosed by the user and other related things (e.g., related machine learning models, social networks, etc.). In some literature, AIAs is also called a model inversion attacks. We do not think this is a reasonable way, and we will explain it later.

\textbf{Example:} Frederickson et al. \cite{fredrikson2014privacy} demonstrated that an attacker can use partial information of an individual's medical records and infer the individual's genotype by using a model trained on similar medical records. In addition, AIAs has been proven to be extremely easy to implement in online social networks, where an attacker can easily infer personal attributes,  such as a citizen's sexual orientation, political orientation, gender, etc. \cite{gong2016you,gong2018attribute,cai2016collective}, through massive amounts of data. Cambridge Analytica even used the method to implement targeted advertising, affecting the US presidential election \cite{isaak2018user}.

\item Model Inversion Attacks (MIs):

\textbf{Definition:} Given a model $F$, the attacker infers information of a specific (class) training data based on the output $Y$ and some auxiliary information (such as the statistics of the training set, or an dataset with the same distribution as the training set, etc., but doesn't have to). MIs initially focused on attacking tabular data, but with its development, today's MIs is no longer limited to low-dimensional data, but is more focused on high-dimensional and complex data, such as images.

\textbf{Example:} A crucial related application of MIs involves the identification of face images, since facial recognition is a commonly used biometric method \cite{galbally2014biometric,rathgeb2011survey}. Once attackers have successfully reconstructed face images of individuals in the private training set, they can use the stolen identities to break into the relevant security systems \cite{wang2021variational}.
\end{itemize}

\noindent\emph{\textbf{3.1.2 Attacks in prediction phase}}
\begin{itemize}
	
	\item Model Inference Attacks:
	
	\textbf{Definition:} It is also known as Model Extraction Attacks (MEIs). Unlike other inference attacks, the goal of MEIs is not to obtain private information about users/training set, but the privacy of released cloud models. Given a trained model $F$, the attacker can extract the parameters, abilities, structure, and other $F$'s information through MEIs without the permission of the model's owner.
	
	\textbf{Example:} Models with commercial value are built at a high cost (time, money, etc.) and can be regarded as a business asset. In MLaaS, users only have access to the released model interface of cloud servers, and cannot access the model's detailed information. Thus, the model's privacy is generally guaranteed. Nevertheless, Tramer \emph{et al.} demonstrated that cloud model's outputs can help attackers steal the capabilities of BigML and Amazon Machine Learning's online models \cite{tramer2016stealing}.
	
\end{itemize}

\subsection{Taxonomy 3MP}
\noindent\emph{\textbf{3.2.1 Flaws in Existing Taxonomy}}

The major problem with the existing taxonomy of inference attacks is that the definitions are not uniform. For instance, in \cite{parisot2021property}, the authors claim membership inference attacks (MIAs) and property inference attacks (PIAs) as a subclass of model inversion attacks (MIs); \cite{hu2021membership} refers to attribute inference attacks (AIAs) as a type of model inversion attack; \cite{zhao2021feasibility} reckon  attribute inference attacks (AIAs) as model inversion attacks (MIs); In \cite{mehnaz2022your,jayaraman2019evaluating}, attribute inference attacks (AIAs) is called model inversion attribute inference attack (MIAIs); \cite{usynin2022beyond} calls model inversion attack (MI) reconstruction attacks (RAs); \cite{ma2021nosnoop} combines property inference attacks (PIAs) and attribute inference attacks (AIA) into one category, collectively called property inference attacks (PIAs); He \emph{et al.} \cite{he2020attacking} call the membership inference attacks (MIAs) a special attribute inference attacks (AIAs). In \cite{bai2021survey,wang2021variational}, inference attacks do not include all of the above five types (we believe that a precise taxonomy is necessary, or it will bring misunderstanding or confusion to subsequent researchers). The property inference attacks (PIAs) defined by Melis \emph{et al.} \cite{melis2019exploiting} is called attribute inference attacks (AIAs) in \cite{gong2018attribute}. Luo \emph{et al.} \cite{luo2021feature} refer to model inversion attacks (MIs) as feature inference attacks. 

We can effortlessly find that the confusing naming system mainly occurred in the past two years' literature, which is unfavourable for subsequent researchers to conduct systematic and in-depth research in this field. It is worth noting that we do not mean that the views of the existing literature are wrong, but it is necessary to highlight a reasonable and uniform naming system.

\noindent\emph{\textbf{3.2.2 Proposed $3$MP}}

Next, we combine the literal definition of "inference" and the development of various inference attacks in recent years to propose a more systematic and reasonable taxonomy. 

Combining Wikipedia and Oxford Dictionaries, "inference" can be defined as "using known information to infer unknown information". Suppose the above $5$ types of inference attacks are all carried out in a white-box \cite{hayes2019logan} or grey-box \cite{zanella2021grey,truex2019demystifying} scenario, then the attacker will have some "known information", which means that these attacks satisfy the definition of "inference" and can be regarded as inference attacks. In a black-box scenario \cite{shokri2017membership}, all types of attacks only have access to the interface of the target model, and there is no obvious "known information" for attackers. However, attackers can use the access right to indirectly obtain the "known information", such as the outputs of the target model, the mapping relationship between inputs and outputs, etc. In addition, after cloud models are released, service providers will generally provide instruction manuals of these models to public \cite{wang2019npufort,melis2019exploiting}. Attackers can use these indirect information to infer "unknown information (users' privacy)". These facts all indicate that the above five inference attacks all meet the definition of "inference" and are in line with the researchers' intuition about inference attacks.

In conclusion, according to the above illumination, we believe that the five attacks MIAs, PIAs, AIAs, MIs, and MEAs should be included in inference attacks and should be listed as much as possible in the subsequent researchers' paper to help the follow-up researchers understand inference attacks more comprehensively.

\begin{figure}[h]
	\centering
	\includegraphics[width=\linewidth]{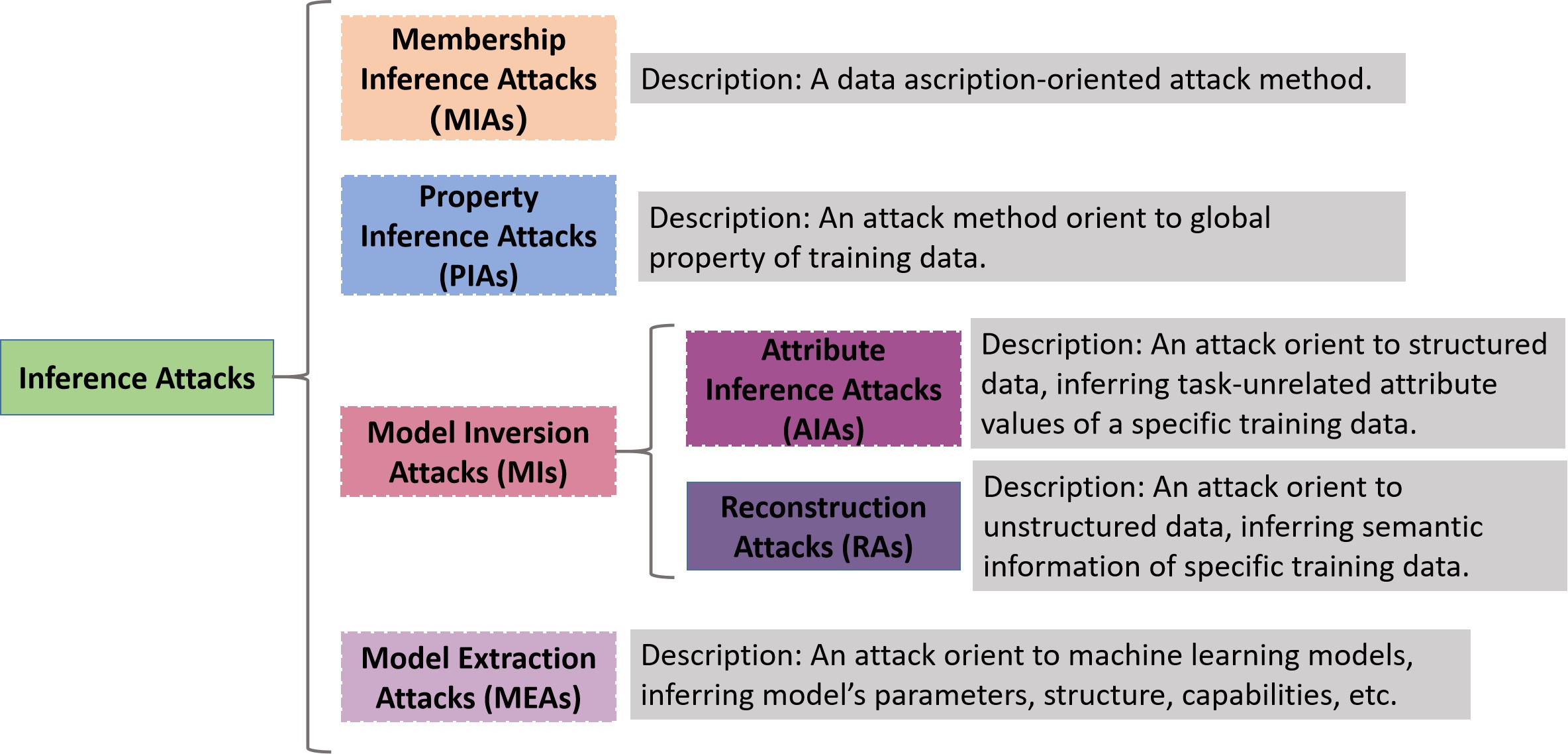}
	\caption{Overview of taxonomy $3$MP.}
	\label{fig4}
\end{figure}

After defining inference attacks, we also need to define each attack clearly in conjunction with the development of inference attacks to avoid confusion in the naming system. To this end, we propose the $3$MP taxonomy, as shown in Figure \ref{fig4}. We suggest dividing inference attacks into four categories, including membership inference attacks (MIAs), model extraction attacks (MEAs), model inversion attacks (MIs), and property inference attacks (PIAs), which allow different types of attacks to have a clear definition that do not overlap with other attacks as much as possible. 

According to the definitions we mentioned in Section $3.1.1$, as two conceptually independent research hotspots, MIAs and MEAs can easily be regarded as secondary classes of inference attacks, which is also in line with research intuition and previous researches.
PIAs and AIAs have similar names, but their attack purposes are quite different, thus it is not reasonable to classify them as one. In addition, the definition of PIAs is relatively independent, and we also suggest it as a secondary class of inference attacks. We found that in the existing research, the definition of MIs is very similar to that of AIAs \cite{fredrikson2014privacy,melis2019exploiting}. The early research of MIAs mainly focused on tabular data. During this period, we can even consider MIAs to be AIAs. However, with the deepening of research, MIAs has gradually expanded to attack high-dimensional data (e.g., images and texts) and has become the focus of researchers. Meanwhile, MIAs is called RAs in some literature. Still, Generative Adversarial Network (GAN) \cite{https://doi.org/10.48550/arxiv.1406.2661}, as essential technical support for RAs, is not satisfactory for the generation of structured data \cite{kehui2019generative,li2021improving,xu2020synthesizing,xu2019modeling,xu2018synthesizing}. Furthermore, from the data flow perspective, whether AIAs or RAs, their information flow is from the model to the training set, which fits very well with the definition and intuition of MIs. Therefore, we can divide MIs into two categories: $1)$ AIAs: a type of model inversion attacks for structured/low-dimensional data; $2)$ RAs: a type of model inversion attacks for unstructured/high-dimensional data.

The proposed taxonomy $3$MP merely changes the existing definition of each attack and is primarily just a redefinition of the inference attack family structure. Therefore, $3$MP will not cause any trouble to current researchers but can provide more rational and clear structured cognition to subsequent researchers. 

\section{Inference attacks in MLaaS}
In this section, we present each type of inference attack according to the $3$MP, including their implementation methods and causes, etc. Due to the different prior knowledge and capabilities possessed by attackers in different scenarios, the effectiveness of various attack's methods is different. Therefore, before presenting various inference attack implementations, we first present the prior knowledge and capabilities that the attackers may holds. Besides, this can help readers to better understand the terminology in inference attacks.

\noindent\emph{\textbf{$\bullet$ Prior Knowledge}}

The prior knowledge an attacker holds when designing an inference attack method largely determines the threat level of the attack. In different scenarios, the prior knowledge possessed by attackers is not the same. For instance, in the federated learning system, the cloud service provider can access the detailed information of the global model as an attacker; Malicious clients can know the distribution of training data in a collaborative learning system; Some attackers only have access to model interfaces, etc. Generally speaking, an attacker with limited knowledge is more challenging to carry out an inference attack, but the damage caused by the attack will be more significant, and vice versa. Depending on the amount of knowledge, attacks can be roughly divided into the following categories.

1) White-box Attacks: Under this setting, the attacker can obtain all the information needed for the attack, e.g., the parameters of the target model \cite{chen2020gan}, the models' type and platform \cite{salem2018ml,truex2019demystifying}, the training set's distribution, and so on. It's worth noting, though, that attackers don't necessarily leverage all of this information. For example, in \cite{nasr2019comprehensive}, the white-box attack performed only used information related to the target model, while the work of Hayes \emph{et al.} \cite{ hayes2017logan} utilised both model information and training set information.

2) Grey-box Attacks: In this setting, the attacker generally does not directly have detailed information about the target model or training set, but can obtain some relatively marginal and limited information, such as the confidence of the model's outputs and the marginal distribution of the training data. In some literature, grey-box attacks can also sometimes be regarded as black-box attacks, which is inconclusive. As in \cite{wu2022exploring}, the confidence-based grey-box attack is also called the black-box attack in \cite{jia2019memguard}.

3) Black-box Attacks: This is the most restrictive attack setting, and the attacker only has access to the model interface and cannot get any other information that would help the attack. For example, in \cite{li2020label,choquette2021label}, the author successfully implemented the attack using only the hard labels output by cloud models, which means that almost anyone can perform inference attacks on cloud models. Therefore, once a black-box attack is successfully implemented in a practical scenario, the harm caused will be very serious.

\noindent\emph{\textbf{$\bullet$ Adversarial Capability}}

In addition to possessing knowledge, it is also possible that an attacker possesses some capability to manipulate the machine learning process or to enhance the effectiveness of the attack. For example, in collaborative learning systems such as federated learning, the malicious participant can control the direction of model training to a certain extent \cite{hitaj2017deep}, and the aggregator can implement attacks based on the differences in the uploaded local models of different participants. The attacks that attackers intervene in the model training process to assist inference attacks are called active attacks, and vice versa are called passive attacks \cite{melis2019exploiting}.

In addition, most inference attacks are query-based, hence different query capabilities also affect the attack effectiveness, such as how many times the target model can be accessed per day, the maximum amount of data that can be uploaded per query, etc. Generally speaking, query ability is one of the cardinal reasons why most inference attacks are challenging to implement in practice, since cloud service providers are usually sensitive to anomalous queries, thus they limit attackers' query number per day.

%攻击者的知识，不同知识下的攻击方法，不同类型模型所面临的MIA，
\subsection{Membership inference attacks (MIAs)}

\noindent\emph{\textbf{4.1.1 Methods of Membership Inference Attacks}}

Machine learning models usually have a large amount of parameters and traverse the training data many times (tens to hundreds) during the training phase, which means that the model can memorize the information of the training set during the training process \cite{carlini2019secret,song2017machine}. Furthermore, models often overfit the training set (the training accuracy is higher than the test accuracy) \cite{murakonda2020ml,zhang2021understanding,yeom2018privacy}. These factors cause a trained machine learning model to behave differently in the face of data in the training and non-training sets. For instance, a cat and dog classifier can correctly classify cats in its training set with high confidence, and relatively low confidence in non-training cats. Attackers can use this behavior to implement MIAs to determine whether the data in their hands is a member of the training set. In $2008$, homer \emph{et al.} \cite{homer2008resolving} used mixed DNA sequences to successfully infer a specific DNA sequence, which became the bud of MIAs in computer science. Based on their ideas, Shokri \emph{et al.} \cite{shokri2017membership} first proposed a member inference attack in the computer field. We have already presented the definition and cases of MIAs in Section \ref{sec3.1}, thus we do not repeat them in this section. Generally speaking, MIAs can be roughly divided into two categories: $1$) Binary classification-based attacks. $2$) Metric-based attacks.

\emph{$1$) Binary classification-based attacks.} Essentially, binary classification-based MIAs are trying to utilize a machine learning model (attack model) to find the difference in the behavior of target model on members and non-members. The shadow training method proposed by Shokri \emph{et al.} \cite{shokri2017membership} is one of the most widely used attack techniques. Specifically, the main idea is that the attacker creates and trains multiple "shadow models" to simulate the behavior of the target model (i.e., if the target model is a facial classifier, then these "shadow models" are also facial classifiers), and then labels the outputs of these "shadow models" (e.g., label $1$ means that the input corresponding to this output is from the training set, and vice versa). Finally,  based on the labeled output, attackers train a binary classifier that can identify whether the input corresponding to the model output is a member. However, training such a binary classifier is challenging.

\begin{figure}[h]
	\centering
	\includegraphics[width=\linewidth]{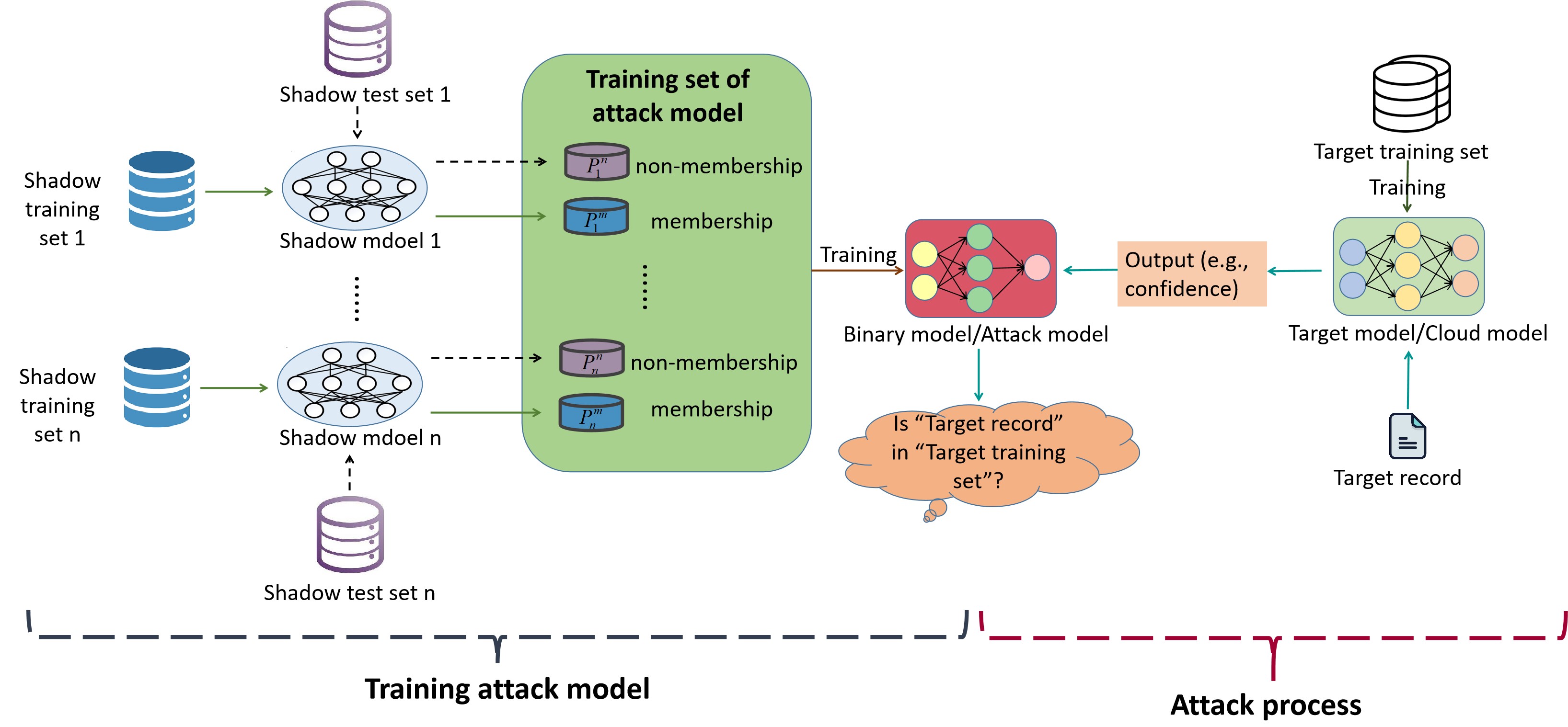}
	\caption{Overview of binary classification-based MIAs.}
	\label{fig5}
\end{figure}

Figure \ref{fig5} shows a typical binary classification-based MIAs paradigm, where "shadow training sets" are used for shadow models' training. Generally, these "shadow datasets" (i.e., shadow test sets and shadow training sets) are disjoint from the target training set. $P^n$ and $P^m$ represent the shadow model's prediction results (such as probability distribution) on the shadow test sets and shadow training sets, respectively. Then use the dataset $\{[(P^n_1,non-membership),(P^m_1,membership)],...,[(P^n_n,non-membership),(P^m_n,membership)]\}$ to train a binary classification model as an attack model. When attackers want to check whether a specific data (target record) they own is in the target training set. They can input the record into the target model to obtain the corresponding output, and then use the attack model to predict the output to complete MIAs. Generally, the more shadow models, the stronger attack effectiveness, because more shadow models can provide more diverse training data for the attack model to better extract the behavioral modes of the target model \cite{shokri2017membership,li2020label,choquette2021label,li2021membership}.

However, the shadow training set is the key of an effective attack model. There are roughly $4$ methods for constructing the shadow training set, as the following \cite{nasr2019comprehensive,lyu2020threats,shokri2017membership,hidano2021transmia}. 
\begin{itemize}
	\item Different data with the same/approximate distribution. For instance, in federated learning, each client needs to train a powerful machine learning model jointly, thus, a malicious client has a dataset that is the same/approximately distributed as other normal clients. These datasets can be directly used as shadow datasets.
	
	\item Model-based synthesis. When attackers do not have any statistical information about the training set of target models, they can synthesize the shadow training set based on the output of target models. The method relies on the intuition that the data classified with high confidence by target model is statistically similar to the target training set. In terms of implementation, the possible data space can be searched through the hill-climbing algorithm through gradually improve the confidence of the synthetic data on target models. 
	
	\item Statistics-based synthesis. Attackers may know marginal distributions of the training set of target models, and they can synthesize shadow training data by sampling these marginal distributions.
	
	\item Noisy real data. An attacker may gain access to a noisy version of the target model's training set. For example, in a location dataset, the shadow training set may be obtained by inverting the attribute values of some data point in the target models' training set.
\end{itemize}

It is worthy note that the attack model and shadow models are built using common machine learning techniques (e.g. CNN, FCNN). Thus, in theory, both can be designed using any state-of-the-art machine learning model/technique.

\emph{$2$) Metric-based attacks.} Compared with MIAs binary classification-based models, metric-based MIAs do not need to train any shadow model or attack model, thus it is much simpler computationally. Generally, metric-based MIAs first perform a certain metric operation on the target model's outputs, and then compare the metric value with a pre-defined threshold to determine whether the target data corresponding to the outputs belongs to the target training set. We can roughly categorize metric-based MIAs into $5$ categories.

\begin{itemize}
	\item MIAs based on prediction correctness \cite{yeom2018privacy}. When using partial information of the target data $X$ for prediction, the target model can still predict it correctly, then $X$ is a member, otherwise it is not. The intuition is that the target model has a good generalization on its training set, hence it is possible to correctly predict it based on partial information of its training data.
	
	\item MIAs based on prediction loss \cite{yeom2018privacy}. When the predicted loss of $X$ returned by the target model is lower th an a predetermined threshold, then $X$ is a member, otherwise it is not. The intuition is that the target model will continuously reduce the prediction loss during the training process to improve its performance, which makes the prediction loss on the training set extremely low.
	
	\item MIAs based on prediction confidence \cite{salem2018ml}. When the target model correctly predicts $X$ and the returned maximum confidence is higher than a predetermined threshold, then $X$ is a member, otherwise it is not. The intuition is that machine learning tends to fit the one-hot encoding of the training data labels during training, hence ideally, the model's prediction confidence on the training set will be infinitely close to $1$.
	
	\item MIAs based on prediction entropy \cite{salem2018ml,song2021systematic}. When the target model correctly predicts $X$, and the prediction entropy is less than a predetermined threshold, then $X$ is a member, otherwise it is not. The intuition is that the generalization of machine learning on its test set is lower than its training set, thus the corresponding prediction entropy on its test set should be greater than its training set. The prediction entropy is defined as follows,
	\begin{equation}
		PE(\hat{p}(y \mid x), y)=-\left(1-p_{y}\right) \log \left(p_{y}\right)-\sum_{i \neq y} p_{i} \log \left(1-p_{i}\right),
	\end{equation}
	where $\hat{p}(y \mid x)$ denotes the confidence corresponding to target data $x$ output  by target model; $p_{i}$ is elements in $\hat{p}(y \mid x)$.
	
	\item MIAs based on adversarial perturbation \cite{del2022leveraging}. When the target model can correctly predict the adversarial sample version of the target data $X$, and the minimal amount of adversarial perturbation is greater than a pre-defined threshold, $X$ is a member, otherwise it is not. The intuition of this attack is that since the machine learning model is more robust to its training set, the minimal amount of perturbation required to transform the training data into an adversarial example is usually higher than to transform the non-training data.
\end{itemize}

\noindent\emph{\textbf{4.1.2 MIAs against Federated Learning}}

Federated learning (FL) is one of the most important privacy-preserving architectures in MLaaS \cite{mcmahan2017communication,mcmahan2017learning}, thus it is necessary to illustrate the impact of MIAs on it. In FL, participants can collaboratively train an efficient machine learning model without exposing their local data, but merely interacting with a cloud server for gradients or parameters. However, this training architecture makes it possible for both participants and cloud servers in FL to be the performers of MIAs. 

Nasr \emph{et al.} \cite{nasr2019comprehensive} believe that an attacker in FL can actively influence the global training process to carry out attacks. They proposed the gradient ascent algorithm (GAA) to increase the loss of the target data. If the target data is a member of other participants' training set, they will quickly reduce the gradients on the target data in the next round of training. The inference accuracy of GAA is $76.7\%$ and $82.1\%$ performed on the participants and the server, respectively.

Melis \emph{et al.} \cite{melis2019exploiting} pointed out that when training models with discrete and sparse data, only the place where data points exist can generate gradients. For example, a model trained with natural languages will only calculate the gradients where the word vector is not $0$ when performing the back-propagation algorithm. The word vectors' value is $0$ means that no word corresponding to the word vector in the current input. A malicious cloud in FL can use this method to directly obtain the training data of each participant. Truex \emph{et al.} \cite{truex2019demystifying} believe that the training data of the participants in federated learning has a clear decision boundary (such as non-IID distribution). Therefore, a insider attacker can judge whether a participant has certain data according to the decision boundary.

Most MIAs in FL focus on inferring whether the target data exists in the training set of the global model. In contrast, HU \emph{et al.} \cite{hu2021source} argue that existing MIAs ignore the source of the training data, i.e. the target data belongs to which participant. Thus, they proposed a source inference attack. They believe that exploring source inference attacks is necessary for privacy protection research, because most FL systems add anonymity mechanisms to protect user privacy (such as scrambled models). For example, with multiple hospitals co-training a Covid-$19$ diagnostic model, MIAs can only reveal who has been tested for Covid-19, but further identifying the source hospital these people come from will make them more vulnerable to discrimination, especially if the hospital is located in a high-risk area or country \cite{devakumar2020racism}. Source inference attack exploits the difference between the performance of a participant's local model on the local training set and of other participants' training set. Hence malicious cloud servers can implement source inference attacks through the prediction loss of the local model.

\noindent\emph{\textbf{4.1.3 MIAs on Machine Learning Models}}

The main and initial research object of member inference attack is classification models. However, with the development of machine learning technology, MIAs for other types of models cannot be ignored. In this section, we select some representative papers for presentation, as shown in Table 1.

% Please add the following required packages to your document preamble:
% \usepackage{multirow}
% Please add the following required packages to your document preamble:
% \usepackage{multirow}
\begin{table}[]
	\tiny
	\centering
	\caption{Partial important studies of MIAs.}
	\begin{tabularx}{\linewidth}{|X<{\centering}|X<{\centering}|X<{\centering}|X<{\centering}|X<{\centering}|X<{\centering}|X<{\centering}|}
		\hline
		Year                  & Literature & Target Model & Knowledge & Attack Techonology & Alogrithm Level & Related Domain \\ \hline
		\multirow{3}{*}{2018} & Long \emph{et al.} \cite{long2018understanding} & Binary/Multi-class Classifier & Black-box  &  Hypothesis Test& Centralized&Image Classification\\ \cline{2-7} 
		
		&Yeom \emph{et al.} \cite{yeom2018privacy} &Multi-class Classifier&Black-box &Prediction Correctness/Loss&Centralized & Image Classification  \\ \cline{2-7} 
		
		& &  &  &  &  &  \\ \hline
		\multirow{4}{*}{2019} & Salem \emph{et al.} \cite{salem2018ml}& Binary/Multi-class Classifier &Black-box &Shadow Training, Prediction Confidence/Entropy  &Centralized &Image/Text Classification              \\ \cline{2-7} 
		
		&Hayes \emph{et al.} \cite{hayes2019logan}&GANs&White/Black-box&Reference Discriminator&Centralized& Image Generation  \\ \cline{2-7} 
		
		&Hilprecht \emph{et al.} \cite{hilprecht2019monte}&GANs/VAEs&White-box&Monte-carlo Integration, Reconstruction & Centralized& Image Generation\\ \cline{2-7} 
		
		&Meli \emph{et al.} \cite{melis2019exploiting}&Multi-class Classifier&White-box&Non-zero Gradients  & Federated& Image/Text Classification\\ \cline{2-7} 
		
		&Nasr \emph{et al.} \cite{nasr2019comprehensive}&Multi-class Classifier&White-box&Shadow Training & Federated& Image Classification\\ \cline{2-7} 
		
		&Sablayrolles \emph{et al.} \cite{sablayrolles2019white}&Multi-class Classifier&Black-box&Prediction Correctness/Loss & Centralized& Image Classification\\ \cline{2-7} 
		&            &              &           &                    &                 &                \\ \hline
		\multirow{3}{*}{2020} &Chen \emph{et al.} \cite{chen2020gan}&GANs/VAEs&White-box&Reference GAN & Centralized& Image Generation\\ \cline{2-7}
		
		&Jayaraman \emph{et al.} \cite{jayaraman2020revisiting}&Multi-class Classifier&Black-box&Shadow Training, Prediction Loss & Centralized& Image Classification\\ \cline{2-7}
		
		&Leino \emph{et al.} \cite{leino2020stolen}&Binary/Multi-class Classifier&White-box&Shadow Model & Centralized& Image Classification\\ \cline{2-7}
		
		&Long \emph{et al.} \cite{long2020pragmatic}&Binary-class Classifier&Black-box&Shadow Training, Hypothesis Test  & Centralized& Image Classification\\ \cline{2-7}
		
		&            &              &           &                    &                 &                \\ \hline
		\multirow{3}{*}{2021} &Chen \emph{et al.} \cite{chen2021machine}&Binary/Multi-class Classifier&Black-box&Shadow Training & Centralized& Image Generation\\ \cline{2-7}
		
		&Choquette \emph{et al.} \cite{choquette2021label}&Multi-class Classifier&Black-box&Prediction Correctness, Adversarial Perturbation & Centralized& Image Classification\\ \cline{2-7}
		
		&Hui \emph{et al.} \cite{hui2021practical}&Binary/Multi-class Classifier&Black-box&Differential Comparisons & Centralized& Image Classification\\ \cline{2-7}
		
		&Li \emph{et al.} \cite{li2021membership}&Multi-class Classifier&Black-box&Adversarial Perturbation & Centralized& Image Classification\\ \cline{2-7}
		
		&Shokri \emph{et al.} \cite{shokri2021privacy}&Binary/Multi-class Classifier&Black-box&Shadow Training& Centralized& Image Classification\\ \cline{2-7}
		
		&Song \emph{et al.} \cite{song2021systematic}&Multi-class Classifier&Black-box&Prediction Entropy & Centralized& Image Classification\\ \cline{2-7}
		
		&Zhang \emph{et al.} \cite{zhang2021membership}&Recommendation Model&Black-box&Shadow Training & Centralized& Recommender System\\ \cline{2-7}
		
		&Shah \emph{et al.} \cite{shah2021evaluating}&ASR Model&Black-box&Shadow Training, Prediction Loss/Correction & Centralized& Speech Recognition\\ \cline{2-7}
		
		&Wang \emph{et al.} \cite{wang2021membership}&KGE model&Black-box&Shadow Training & Centralized&Knowledge Graph\\ \cline{2-7}
		
		&Shafran \emph{et al.} \cite{shafran2021reconstruction}&Image Translation Model, Semantic Segmentation Model&Black-box&Reconstruction Error & Centralized& Image Segmentation\\ \cline{2-7}
		&            &              &           &                    &                 &                \\ \hline
		
		\multirow{3}{*}{2022} &Carlini \emph{et al.} \cite{carlini2022membership}&Multi-class Classifier&White/Black-box&Shadow Training & Centralized& Image/Text Classification\\ \cline{2-7}
		
		&Shi \emph{et al.} \cite{shi2022membership}&Wireless Signal classifier&Black-box&Shadow Training, Prediction Confidence & Centralized& Wireless Signal Classification\\ \cline{2-7}
		
		&Conti \emph{et al.} \cite{conti2022label}&Node Classifier&Black-box&Shadow Training & Centralized& Graph Classification\\ \cline{2-7}
		
		&Li \emph{et al.} \cite{li2022user}&Metric Embedding Model&Black-box&Shadow Training & Centralized& Person Re-identification\\ \cline{2-7}
		
		&He \emph{et al.} \cite{he2022semi}&Semi-supervised Model&Black-box&Shadow Training & Centralized& Image Classification\\ \cline{2-7}
		
		&Wu \emph{et al.} \cite{wu2022membership}&Text-to-image Generation Model&Black-box&Generation Quality, Reconstruction Loss, Semantic Reflection& Centralized& Image Generation\\ \cline{2-7}
		
		&Liu \emph{et al.} \cite{liu2022membership}&Binary/Multi-class Classifier&Black-box&  Jacobian matrix, Perturbation & Centralized& Image/Sequence Classification\\ \cline{2-7}
		
		&Yuan \emph{et al.} \cite{yuan2022membership}& Pruned Model&Black/White-box&Prediction Sensitivity/Confidence & Centralized& Image/Sequence Classification\\ \cline{2-7}
		
		&Suri \emph{et al.} \cite{suri2022subject}&Multi-class Classifier&Black-box& Prediction Loss, Loss-Across-Rounds  & Federated& Image Classification\\ \cline{2-7}
		&            &              &           &                    &                 &                \\ \hline
	\end{tabularx}
	\begin{tablenotes} 
	\item $\cdot$ To avoid unnecessary misunderstanding, in this table, we classify grey-box attacks into black-box attacks.
\end{tablenotes} 
\end{table}

\noindent\emph{\textbf{(1) Classification Models}}

An attack against a classification model was first proposed by Shokri \emph{et al.} \cite{shokri2017membership}, who used multiple shadow models to build a training set for the attack model. Salem \emph{et al.} \cite{salem2018ml} extended the study of Shokri \emph{et al.}, proving that the distribution of the shadow training set is not necessarily the same as the target training set, and that only one shadow model can have comparable attack performance. Furthermore, they also propose metric-based attack methods that exploit confidence and prediction entropy. Long \emph{et al.} \cite{long2018understanding,long2020pragmatic} implemented MIAs for machine learning models without overfitting. They argue that even if a model generalizes very well, some data in its training set can still uniquely impact the model. They prove that when the error between the training accuracy and test accuracy of the model is less than $1\%$, the membership of some training data can still be inferred by using this impact.

The above attacks assume that the attacker can obtain detailed information (such as confidence) about the target model's outputs. Li \emph{et al.} \cite{li2021membership} demonstrated that an attacker only needs to know the hard labels to implement MIAs. They proposed transfer-based MIA and adversarial perturbation-based MIA. Transfer-based MIA utilizes a shadow model to simulate the capability of the target model. In theory, when the shadow model is similar enough to the target model, their output to the target data will be similar enough. The adversarial perturbation-based MIA transforms the target data into adversarial samples by adding adversarial perturbations to it, and measures the amount of perturbation to judge whether the target data is a member. Choquette \emph{et al.} \cite{choquette2021label} proposed data augmentation-based MIA, and the idea is similar to MIA adversarial perturbation-based MIA. That is, if a data is a member, the target model will still generalize well to its augmented version.

MIA in white-box scenarios was first proposed by Nasr \emph{et al.} \cite{nasr2019comprehensive}, who exploited the intermediate outputs of the target model to improve attack effectiveness, such as gradients. Leino \emph{et al.} \cite{leino2020stolen} argue that the attack proposed by Nasr et al. has too strong assumptions, such as the attacker knows most of the training set of the target model. Therefore, they approximated each neural network layer as a local linear model and used the Bayes-optimal attack to compute the final membership decision.

\noindent\emph{\textbf{(2) Generative Models}}	

Hayes \emph{et al.} \cite{hayes2019logan} implemented MIAs for GANs for the first time in both white-box and black-box scenarios. In the white-box scenario, the attacker inputs all target data to the discriminator and selects the top $50\%$ of the samples as members. In the black-box scenario, the attacker collects the samples generated by the generator to train a classifier to simulate the discriminator. After the classifier is trained, the attack process of the white-box scenario is executed. Hilprecht \emph{et al.} \cite{hilprecht2019monte} generate data that approximates the target sample in the sample space. Then use the Monte Carlo integration attack to approximate the probability that the record is a member. The intuition of this attack is that the generator of a GAN should be able to generate data close to the target sample in the sample space.

Chen \emph{et al.} \cite{chen2020gan} proposed MIAs for attacking GANs in both white-box and black-box settings. Similar to the work of Hilprecht \emph{et al.}, they reconstruct synthetic data close to the target data through the target generator, and evaluate the distance between the synthetic data and the target sample. To make more accurate probability estimates, they trained a reference GAN with related but disjoint datasets to narrow the distance. When the distance error is less than a certain threshold, the target record is a member. This attack can threaten many advanced generative models, such as DCGAN  \cite{radford2015unsupervised}, VAEGAN \cite{larsen2016autoencoding}, MEDGAN \cite{choi2017generating}, etc. Furthermore, it is worth noting that VAEs are more vulnerable than GANs because VAEs are more prone to overfitting.
		
\noindent\emph{\textbf{(3) Embedding Models}}

Embedding technology has been applied in many important fields, such as NLP, social networks, etc. The core of embedding is to map discrete natural information into spatial vectors for computer processing. Song \emph{et al.} \cite{song2020information} first proposed MIAs for word embedding models. They used metric-based MIAs to obtain the similarity scores of sliding windows of target words or sentences and judge their membership status. The first implementation of MIAs for graph embedding models was performed by Duddu \emph{et al.} \cite{duddu2020quantifying}, who designed a binary classification attack model using shadow training techniques. Furthermore, they also demonstrated that confidence scores-based MIAs can also compromise graph embedding networks

\noindent\emph{\textbf{(4) Regression Models}}

MIAs for deep regression networks were first proposed by Gupta \emph{et al.} \cite{gupta2021membership}. They implemented MIAs on a regression model predicting age by MRI in a white-box scenario. They used the target model's gradients, and outputs to train a binary classifier as an attack model, confirming that the regression model is also susceptible to MIAs.

\subsection{Property Inference Attacks (PIAs)}
\noindent\emph{\textbf{4.2.1 Methods of Property Inference Attacks}}

\begin{figure}[h]
	\centering
	\includegraphics[width=0.8\linewidth]{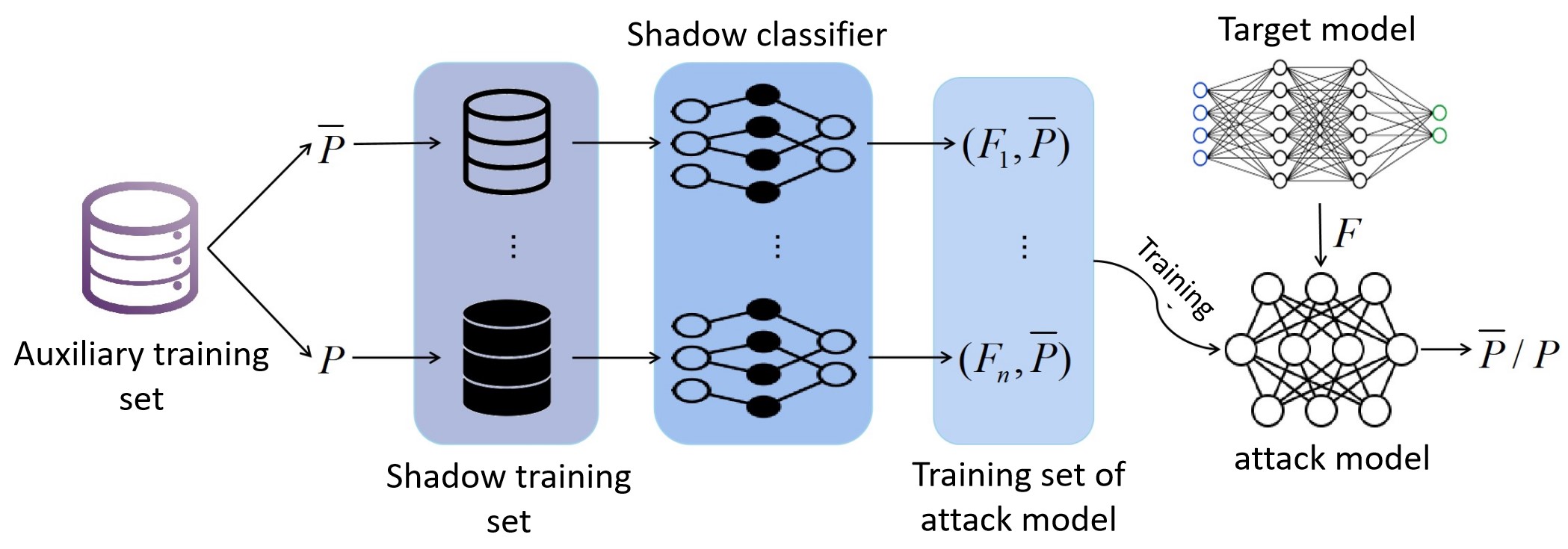}
	\caption{Overview of PIAs.}
	\label{fig6}
\end{figure}

Unlike MIAs, the attackers of PIAs do not hold the target data, and they aim to infer the global statistics of the target training set, such as the ratio of male and female, and the proportion of people of color in the target training set. Intuitively, attackers reckon that models trained with similar datasets and algorithms will have similar internal properties. This similarity can inadvertently expose model-related information. Ateniese \emph{et al.} \cite{ateniese2013hacking} first defined PIAs and designed the first MIA (FPIA) method using shadow training techniques, as shown in Figure \ref{fig6}.

In FPIA, attackers have an training set that is disjoint but identically distributed with the target training set.  They group the auxiliary training set according to whether the data in it has a certain property $P$ to get the shadow training set. The shadow training set is used to train multiple shadow classifiers that function same as the target model. After the training, "the model parameters of each shadow classifier" and "whether the training set has the attribute $P$" form a data pair to obtain the training set of the attack model, and then utilise the training set to train the attack model. Ateniese \emph{et al.} Ateniese \emph{et al.} confirmed that FPIA could successfully attack traditional machine learning models such as support vector machines and hidden Markov chains. They also demonstrated that even record-level differential privacy is difficult to defend against FPIA.

However, FPIA has limited performance in the face of neural networks, which is one of the most widely used machine learning techniques. The researchers believe that this is because the deep neural network has an extremely large number of parameters, which makes the training of shadow models and attack models very difficult. Therefore, in $2018$, Ganju \emph{et al.} \cite{ganju2018property} first proposed a PIA that can attack deep fully connected neural networks (FCNNs). They believe that the permutation invariance\footnote{permutation invariance: For each hidden layer of a trained FCNN, arbitrarily changing the order of the nodes in it can get an FCNN equivalent to the original one.} (PI) of neural nodes of FCNNs is one of the reasons why FPIA cannot implement effective attacks on FCNNs. PI enables FCNNs to have many equivalent networks, and the number of equivalent networks grows geometrically with the number of network nodes. This invariance causes FPIA to have a high computational cost in training shadow models. To this end, Ganju \emph{et al.} designed two methods to solve this problem. The first method is to order the neurons of each hidden layer so that each FCNN has its canonical form. Implementing PFIA with FCNNs' canonical form can reduce computational overhead to a certain extent. Compared with the baseline method, the PI-based method improves the lower and upper bound of inference accuracy by 30 and 27 percentage points, respectively. The second method uses the DeepSet technique \cite{zaheer2017deep} to represent each hidden layer as a set instead of a vector, and train the attack model based on this set. The DeepSet not only captures permutation invariance, but also significantly reduces the number of parameters of the attack model, making the attack easier to carry out.

\noindent\emph{\textbf{4.2.2 PIAs against Federated Learning}}

In 2019, Melis \emph{et al.} \cite{melis2019exploiting} extended PIAs to federated learning for the first time. They point out that malicious participants can implement PIAs using model snapshots at each update. Specifically, they assume that the attacker has an auxiliary dataset and divide it into $D_p$ and $D_{\hat{p}}$ based on whether the data in the dataset has property $p$ or not. In each model update process, attackers use the model snapshot $F$ to obtain the gradients corresponding to $D_p$ and $D_{\hat{p}}$, respectively, and then label them accordingly. Finally, the attack model is trained with the labeled dataset. In the above attacks, the attacker does not interfere with the training of the target model, which is called passive attacks.If attackers interfere with the training of the target model (usually to make the PIA stronger), these attacks are called active attacks. Melis \emph{et al.} locally concatenate and train the attack model with the global model to form a multi-task machine learning model (main task and property classification task), resulting in an active attack version of PIAs.

Neural networks are widely used in many fields because of their outstanding performance (classification, generation, etc.). However, studies have shown that attackers can implement poisoning attacks by modifying local data labels, decreasing the performance of collaborative training models significantly. This process can enhance the effect of PIA. Specifically, Mahloujifar \emph{et al.} \cite{chase2021property} believe that the attacker can treat the target model as a Bayesian optimal classifier and use the poisoning distribution to implement PIA based on the shadow training technique. Their theory proves that the attack is always successful as long as the learning algorithm has good generalization.

\noindent\emph{\textbf{4.2.3 PIAs on Machine Learning Models}}

Research on PIAs initially focused on classification/fully connected models, but there are still many types of models facing the threat of PIAs.

\noindent\emph{\textbf{(1) Generative Models}}

Zhou \emph{et al.} \cite{zhou2021property} first proposed PIA against adversarial generative networks (GANs). Their intuition is that the data generated by GANs can reflect certain properties of the training set. For instance, a GAN trained on white males, its generated images have a high probability of being white males. Therefore, the attacker can process the data generated by the target GAN to obtain the underlying properties of the target dataset.

\begin{table}[]
	\tiny
	\caption{Summaries of partial important PIAs' studies.}
	\centering
	\begin{threeparttable} %添加此处
		\begin{tabularx}{\linewidth}{|X<{\centering}|X<{\centering}|X<{\centering}|X<{\centering}|X<{\centering}|X<{\centering}|X<{\centering}|X<{\centering}|}\hline
			\multirow{2}{*}{\textbf{Literature}} & \multirow{2}{*}{\textbf{AC}} & \multicolumn{2}{c|}{\textbf{Target model}}    & \multirow{2}{*}{\textbf{Technology}} & \multirow{2}{*}{\textbf{Scenario}} & \multirow{2}{*}{\textbf{Metric}} & \multirow{2}{*}{\textbf{DC}} \\ \cline{3-4}
			& & \multicolumn{1}{c|}{\textbf{Type}} & \textbf{Instance} & & & & \\ \hline
			$2013$ Ateniese \emph{et al.} \cite{ateniese2013hacking} & White-box& \multicolumn{1}{c|}{TM}   &   \makecell[c]{SVM\\HMM\\DT}  &  ST   & Released Classifier &     Recall; Precision; Accuracy   & \textcolor{red}{DP} \\ \hline
			
			$2018$ Ganju \emph{et al.} \cite{ganju2018property} & White-box& \multicolumn{1}{c|}{FCNN} & Deep FCNN & \makecell[c]{ST; Sorting\\Deepset} & Released Classifier  & Accuracy   & \textcolor{blue}{NMT; Noise; Encoding} \\ \hline	
			
			$2019$ Wang \emph{et al.} \cite{wang2019property} & White-box& \multicolumn{1}{c|}{FCNN} & Deep FCNN & \makecell[c]{ST; Sorting\\Deepset} & Released Classifier  & Accuracy   & \textcolor{blue}{NMT; Noise; Encoding} \\ \hline
			
			$2019$ Gopinath \emph{et al.} \cite{gopinath2019property} & White-box& \multicolumn{1}{c|}{DNN} & \makecell[c]{Deep FCNN\\CNN} & \makecell[c]{Decision tree;\\ Reluplex \cite{katz2017reluplex}}  & Released Network  & ID; VP   & - \\ \hline	
			
			$2020$ Shen \emph{et al.} \cite{shen2020exploiting} & White-box& \multicolumn{1}{c|}{CNN} & \makecell[c]{Custom CNN} & \makecell[c]{ALM\\ }  & Federated Learning  & Accuracy   & - \\ \hline		
			
			$2021$ Crectu \emph{et al.} \cite{crectu2021dataset} & Black/White-box& \multicolumn{1}{c|}{TM} & \makecell[c]{MLP\\LR}  & ST, CC & Accessible and Non-access Model  & Relevance; Accuracy   & - \\ \hline
			
			$2021$ Parisot \emph{et al.} \cite{parisot2021property} & \makecell[c]{White-box}& \multicolumn{1}{c|}{CNN} & \makecell[c]{VGG16} & \makecell[c]{ST} & Released Classifier  & \makecell[c]{Accuracy}   & - \\ \hline

			$2022$ Zhang \emph{et al.} \cite{zhang2022inference} & \makecell[c]{Black-box}& \multicolumn{1}{c|}{GNN} & \makecell[c]{SAGE \cite{hamilton2017inductive}} & \makecell[c]{Auto-encoder\\Decoder} & Released GNN  & \makecell[c]{Accuracy}   & - \\ \hline
			
			$2022$ Zhou \emph{et al.} \cite{zhou2021property} & \makecell[c]{Black-box}& \multicolumn{1}{c|}{GAN} & \makecell[c]{WGAN\\DCGAN} & \makecell[c]{ST\\Property classifier} & Accessible GAN  & \makecell[c]{AUC\\Accuracy}   & \textcolor{green}{Rebalancing dataset} \\ \hline		
			
			$2022$ Mahloujifar \emph{et al.} \cite{chase2021property} & Black-box& \multicolumn{1}{c|}{FCNN, TM} & \makecell[c]{Deep FCNN\\LR}  & ST, PA & Collaborative Learning  & Precision; Recall   & - \\ \hline	
			
		\end{tabularx}
		\begin{tablenotes} 
			\item \textcolor{green}{Green font}: Effective countermeasure; \textcolor{blue}{Blue font}: Unproven  countermeasure; \textcolor{black}{Black font}: Ineffective  countermeasure;
			
			\item \textbf{AC}: Adversary capabilities;   \textbf{DC}: Discussed countermeasure;  \textbf{TM}: Traditional models; \textbf{ST}: Shadow training; \textbf{CC}: Correlation Coefficient; \textbf{MLP}: Multi-layer Perceptron;  \textbf{DP}: Differential privacy; \textbf{NMT}: Node Multiplicative Transformations; \textbf{ID}: Interval detection; \textbf{VP}: Visual perception; \textbf{LR}: Logistic regression; \textbf{PA}: Poisoning attacks; \textbf{ALM}: Aggregating Local Model
		\end{tablenotes} 
	\end{threeparttable} 
\end{table}

\begin{figure}[h]
	\centering
	\includegraphics[width=0.7\linewidth]{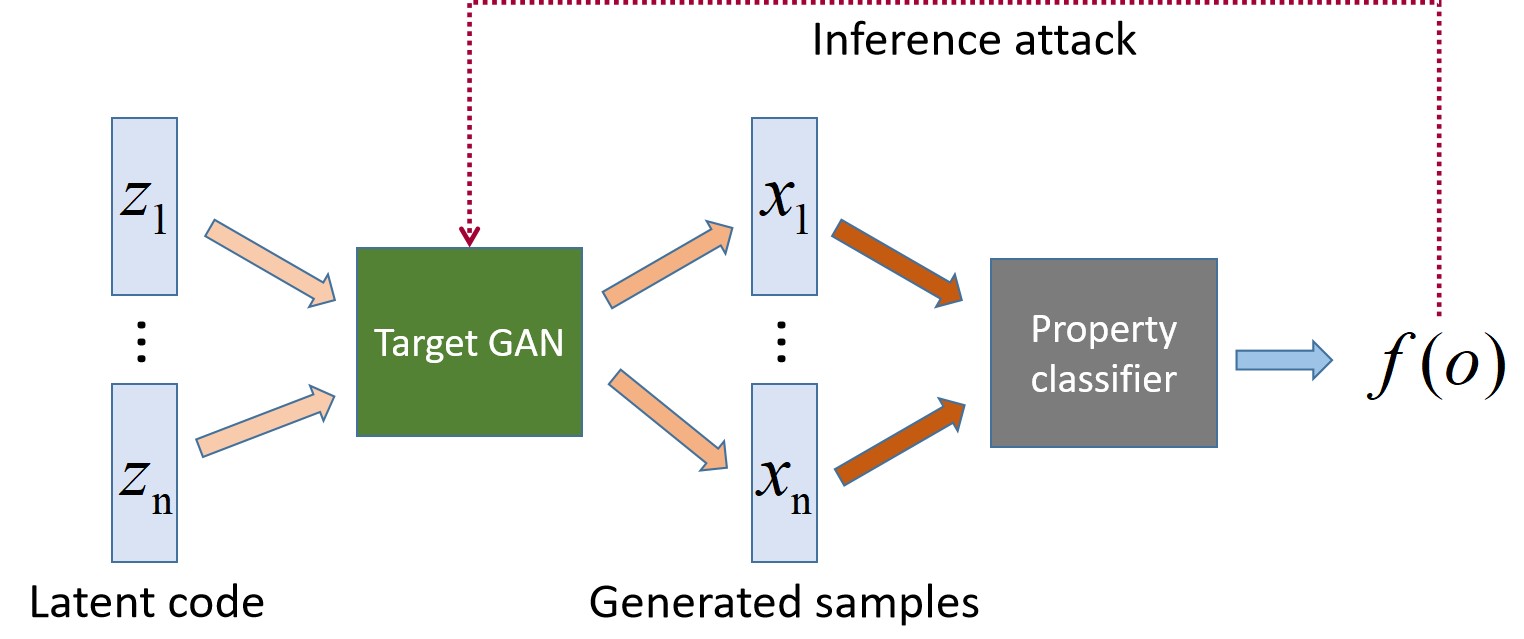}
	\caption{Workflow of the general PIAs strategy for generated models.}
	\label{fig7}
\end{figure}
Figure \ref{fig7} depicts a generalized PIA strategy for GANs. The strategy is mainly divided into $3$ steps. 

\ding{172} The attacker first samples the latent codes $Z={z_1,z_2,...,z_n}$ and then uses them to query the target GAN to obtain the generated samples $X={x_1,x_2,...,x_n}$.  Zhou \emph{et al.} propose different methods to sample $Z$ in black-box and grey-box scenarios, which we present later.

\ding{173} Next, the attacker builds an property classifier to classify $X$. For example, if the attacker wants to infer the gender distribution of the target training set, then the property classifier's task is to classify the gender of $X$. property classifiers are usually trained using an auxiliary training set that distribution is similar but disjoint to the target.

\ding{174} In the end, the attacker designs a function $f$ for calculating the a certain property of target data distribution based on the property classifier's outputs, defined as:

\begin{equation}
	f \left(\left\{M_{\mathcal{Property\_Classifier}}\left(\mathrm{G}_{\text {target }}\left(z_i\right)\right)\right\}_{i=1}^{|\mathbf{X}|}\right)
\end{equation}

\begin{figure}[h]
	\centering
	\includegraphics[width=0.9\linewidth]{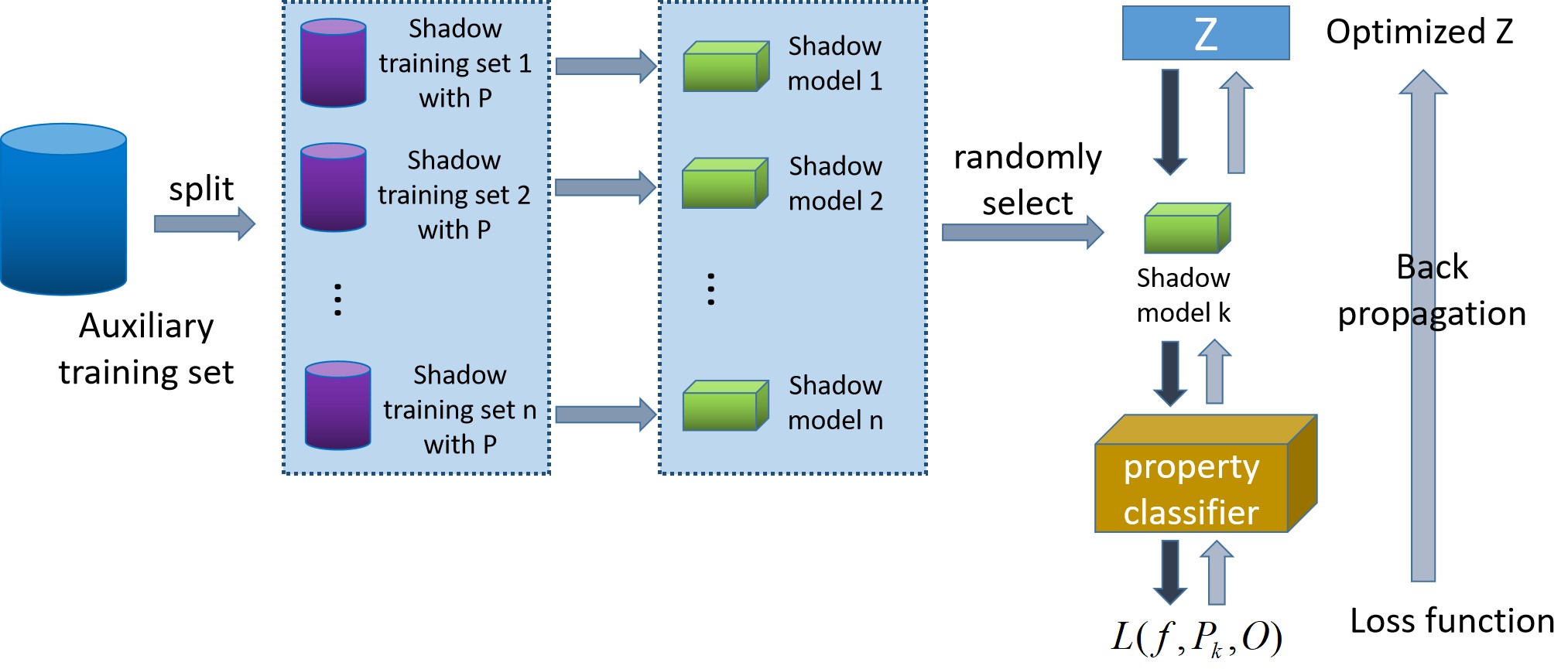}
	\caption{Methodology of optimizing input latent code $Z$.}
	\label{fig8}
\end{figure}

In black-box scenario, the attacker can only randomly obtain the latent code $Z$, such as automatically generated by the victim's system. That is, when attackers queries the target GAN, they only need to send a query request without providing $Z$. The black-box scenario requires the attacker to perform a large number of queries on the target GAN to implement an effective attack, which limits the effectiveness of PIA. In grey-box scenario, the attackers have auxiliary training sets and can also provide $Z$ to the target GAN during the query process. They use shadow training techniques to craft the latent code to make the attack more powerful, as shown in Figure \ref{fig8}.

The attackers first divide the auxiliary training set into $n$ shadow training sets, each filled with all the spatial distributions of a target property $P$. For example, if $P$ stand for gender, the shadow training set can be split into male and female. Then, using these shadow training sets to train $n$ shadow models and randomly select one of them to optimize the latent codes.

\noindent\emph{\textbf{(2) Embedding Models}}

Zhang \emph{et al.} \cite{zhang2022inference} proposed a PIA for graph embedding networks. However, according to $3$MP's definition, their proposed method should be classified as a model inversion attack (MI), since their goal is to infer properties of a specific target model's input rather than statistics of the target training set. To our knowledge, no research has yet designed PIAs for embedded models that conform to the $3$MP definition. Perhaps shadow training techniques can be used to implement PIAs for embedded models, but we have not collected convincing literature. Research in this area has a gap.

If we present PIAs strictly according to the definition of 3MP, then PIAs are a niche study. Therefore, compared with other inference attacks, there is not much literature related to it each year. In the past two years, many publications have easily mixed the concepts of PIAs with MIAs, AIAs or RAs \cite{mo2020layer,pareekproperty,xu2020subject}. We believe that this is not conducive to subsequent in-depth research and the study of new researchers. Besides,

\subsection{Model Inversion Attacks}
According to the definition of 3MP we described in section $3$, we can classify model inversion attacks into two categories, namely attribute inference attacks (AIAs) and reconstruction attacks (RAs). Unlike PIAs and MIAs, model inversion attacks aim to infer specific data information, such as a person's age, image content, etc.

\noindent\emph{\textbf{4.3.1 Attribute Inference Attacks (AIAs)}}

The idea of AIAs originated earlier than other inference attacks, because its initial research field mainly focused on online social network (OSN), a field that was popular before machine learning. User relationships in social networks are rich, which provides the basis for implementation of inference attacks based on information from other users/things. In the early stages of AIAs research, researchers classified AIAs into three main categories.

\emph{$1$) Friend-based attribute inference}

Friend-based attribute inferences mainly take advantage of the homogeneity between data, i.e., you are who you know. The attack mainly consists of two steps: 1) The attacker first collects the friend list of the target user; 2) Then, the attacker infers the attributes of the target through the public information of the target user's friends. For example, if many of the target user's friends have AIDS, it may be considered that the target user is also an AIDS patient. Because people in the same social circle usually have many similar attributes.

He \emph{et al.} \cite{he2006inferring} studied the impact of social relations on private reasoning. For example, to infer a person's age, consider the age of his classmates instead of his colleagues. They used Bayesian networks to model person relationships in social networks, and studied the impact of prior probability, influence strength, and society openness to the inference accuracy on the LiveJournal social network dataset \cite{mackinnon2006age}. However, Bayesian models are difficult to scale in the machine learning community. Lindamood \emph{et al.} \cite{lindamood2009inferring} improved the Bayesian-based method and used other users' attributes (such as company, city, etc.) to conduct AIA. Unfortunately, their attack is ineffective against users who do not share their information and is limited by the Bayesian model.

Zheleva \emph{et al.} \cite{zheleva2009join} have shown how to infer attributes of users using an online social network that mixes public and private profiles. They clarified that a strong attack could be implemented through friend links. Specifically, they represented each user as a binary vector with a value of $1$ if the user was a friend of the person corresponding to a certain feature. Then, these feature vectors are used to learn a classifier for attribute inference. Thomas \emph{et al.} \cite{thomas2010unfriendly} studied the inference of attributes such as gender, political affiliation, and religious opinion using multi-label classifications combining the target user's friends and published posts.

Mislove \emph{et al.} \cite{mislove2010you} identified local communities in a social network by users with the same attributes, and all users in the community have the same attributes. This method cannot infer users outside the local community. For example, Traud \emph{et al.} \cite{traud2012social} found that MIT male communities were associated with their residence, but females were not.

\begin{table}[]
	\tiny
	\caption{Summaries of partial important AIAs' studies.}
	\centering
	\begin{tabularx}{\linewidth}{|X<{\centering}|X<{\centering}|X<{\centering}|X<{\centering}|X<{\centering}|X<{\centering}|X<{\centering}|X<{\centering}|}
		\hline
		Literature & Attacker& Knowledge Level/AIA's type & Attack Model & Scenario &  Countermeasure-discussed &  Countermeasure-proposed & Counter Effectiveness \\ \hline
		%		$2016$ Gong \emph{et al.} \cite{gong2016you} & Any party & Publicly available social structures, user attributes, and behaviors & SBA model & OSN & No & No &      ---         \\ \hline
		
		$2016$ Kumar \emph{et al.} \cite{kumar2016improving} & Any party & Social graph & Closeness centrality, Betweenness  & OSN & No & No &      ---            \\ \hline
		
		$2018$ Gong \emph{et al.} \cite{gong2018attribute} & Any party & Publicly available social structures, behaviors & VIAL-based SBA model  & OSN & Yes & No &      ---            \\ \hline
		
		$2018$ Alipour \emph{et al.} \cite{alipour2019gender} & Any party & Target user's public pictures-related comments and alt-text & NLP model  & OSN & No & No &      ---            \\ \hline
		
		$2018$ Li \emph{et al.} \cite{li2018differentially} & Any party & Community (e.g., movie community) & CDAI; Community detection  & OSN & Yes & Yes & Effective           \\ \hline
		
		$2020$ Pijani \emph{et al.} \cite{pijani2020you} & Any party & Target user's shared images & Comments' Emoji analysis  & OSN & No & No &      ---            \\ \hline
		
		$2021$ Zhang \emph{et al.} \cite{zhang2021tea} & Any party & Social behaviors & TEA-RNN & OSN & No & No &      ---            \\ \hline
		
		$2021$ Pasquini \emph{et al.} \cite{pasquini2021unleashing} &Central server&Black-box&CNN&Split Learning \cite{poirot2019split}& Yes &No&      ---             \\ \hline
		
		$2022$ Cretu \emph{et al.} \cite{cretu2022querysnout} &Any party&Access to the query-based system&QuerySnout&Query-Based Systems& No &No&      ---             \\ \hline
		
		$2022$ Aalmoes \emph{et al.} \cite{aalmoes2022dikaios} &Any party&Access to the target model, a non-overlapping auxiliary dataset&Binary classifier&Cloud service& Yes &No&      ---             \\ \hline
		
	\end{tabularx}
	\begin{tablenotes} 
		\item \textbf{pMRF}: Pairwise markov random field; \textbf{OSN}: Online social network; \textbf{Any party}: The attacker could be OSN provider, advertiser, data broker, or cyber criminal, etc.; \textbf{SBA}: Social-behavior-attribute network model; \textbf{TEA-RNN}: Topic-enhanced attentive RNN; \textbf{VIAL}: Vote distribution attack; \textbf{NLP}: Nature language processing; \textbf{CDAI}: Infer attributes based on community; 
		Detection)
		\item “Access“ means that the attacker has the API access permission of the target model, but does not know the structure and parameters of the target model.
	\end{tablenotes} 
	\label{AIAs}
\end{table}

\emph{$2$) Behavior-based attribute inference}

The core idea of behavior-based attribute inference is: you are how you behave. For example, an attacker can judge a target user's private attributes through him/her behavior, such as liking, following, and joining a group chat. Weinsberg \emph{et al.} \cite{weinsberg2012blurme} used the behavior of users to rate different movies to infer their gender. Specifically, they construct the user's rating behavior as a feature vector. The value of the ith item of this vector is equal to the user's rating for the ith movie, and if the user has not seen this movie, the corresponding value is 0. They confirmed that logistic regression performed better than using SVM and Bayesian modeling. Bhagat \emph{et al.} \cite{bhagat2014recommending} stated that if users were encouraged to leave textual reviews for movies, it would lead to stronger inference attacks. However, this does not quite work in the real world.

Chaabane \emph{et al.} \cite{chaabane2012you} build a corpus using the music that users publicly like, extracting and analyzing imperceptible musical topics. They augmented this corpus using Wikipedia and derived probabilistic models  to compute the belonging of users to each of these topics. Similar works include \cite{weinsberg2012blurme,kosinski2013private,luo2014you,cai2016collective}, etc. Li \emph{et al.} \cite{li2018differentially} et al. considered that users in the same community (such as joining the same movie group) would have similar attributes. Thus they proposed an AIA based on community detection. Furthermore, they used differential privacy to defend the proposed method, and the experiments proved that although it is effective, it will reduce the performance of the recommendation system.

\emph{$3$) Hybrid attribute inference}

The two type of methods above-mentioned are based on either social structure or behaviour, but not both. Therefore, the inference accuracy they can achieve is limited in practical applications. Furthermore, if we consider users' social structure and behaviour simultaneously, the nature of AIAs will change, because the features they represent will show different sparsity and scale. Numerous studies have shown that simply linking these two sources of information can even reduce the success rate of AIA. For instance, in 2018, $2016$ Gong \emph{et al.} \cite{gong2016you} first proposed a social-behavior-attribute (SBA) network model to integrate users' social structure, behavior, and attributes in a unified framework. Then, they designed a vote distribution attack (VIAL) based on the SBA model to perform attribute inference.

\emph{$4$) Other attribute inference}

In addition to the above-mentioned three mainstream methods, there are other methods. Alipour \emph{et al.} \cite{alipour2019gender} crawled the comments related to the pictures posted by the target user, then built an NLP model to analyze these comments, thereby inferring the gender of the target user. Pijani et al.\cite{pijani2020you} pointed out that the gender of the target user can be inferred from the emoticons commented by others on the target user's post. Specifically, the attackers divided emoticons into three categories: 1) Generated alt-text. 2) Emoji.3) Emoticons. Furthermore, record the reactions of Facebook users when they observe different expressions to infer the privacy of target users. 

\begin{figure}[h]
	\centering
	\includegraphics[width=0.7\linewidth]{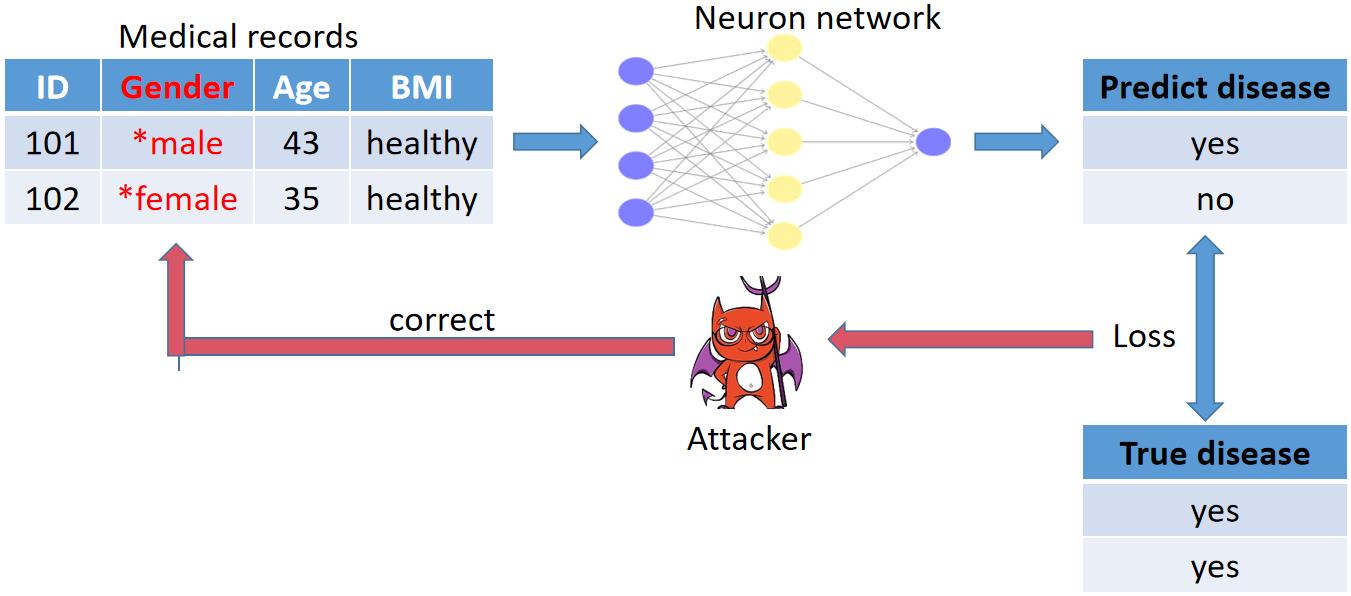}
	\caption{Attribute inference attack with deep neuron network.}recursion
	\label{fig9}
\end{figure}

In Figure \ref{fig9}, the attacker possesses the patient's ID, age, BMI, and other attribute values, but not the patient's gender. The attacker inputs the known patient attributes into the model to obtain the posterior (predicted results). Then combines the posterior to make a loss to correct the missing attribute values, thereby obtaining the attributes that the attacker is interested in.

Zhao \emph{et al.} \cite{zhao2021feasibility} proposed approximate AIA, which is a natural extension of AIAs and is suitable for inferring continuous properties. In approximate AIA, the attacker only needs to approximate the target property by a given distance metric. They showed that approximate AIA is effective in scenarios where AIAs cannot be successfully attacked, and the effect increases as the target model become more overfit. Feng \emph{et al.} \cite{feng2021attribute} used the shadow training technique to conduct an attribute inference attack on a speech emotion recognition system in federated learning for the first time. Their experiments presented that shared model updates in FL may lead to AIA.

We have listed some important attribute inference attacks in Table \ref{AIAs} for readers' reference.

\noindent\emph{\textbf{4.3.2 Reconstruction Attacks (RAs)}}

Based on the $3$PM proposed in this paper, reconstruction attacks are ways to recover unstructured data such as images, audio, and video. According to existing studies, we can classify RAs into three categories depending on the different attack target.

$1$) Reconstruct the sample features belonging to a certain category in the target training set. For example, an attacker can use a target model to recover a person's facial information to bypass facial recognition-based access control systems.

$2$) Reconstruct the data information of the model's inputs. For example, in federated learning, the server does not know the specific information (such as image representation and sentence.) about the client's local training set. However, malicious service providers can leverage RAs to reconstruct the client's training set.

$3$) Reconstruct data labels. For example, in federated learning, a malicious server can reconstruct the label of training data through model parameters or gradients uploaded by clients.

In addition, based on different attack scenarios, we can also divide RAs into three categories.

\begin{figure}[h]
	\centering
	\includegraphics[width=1\linewidth]{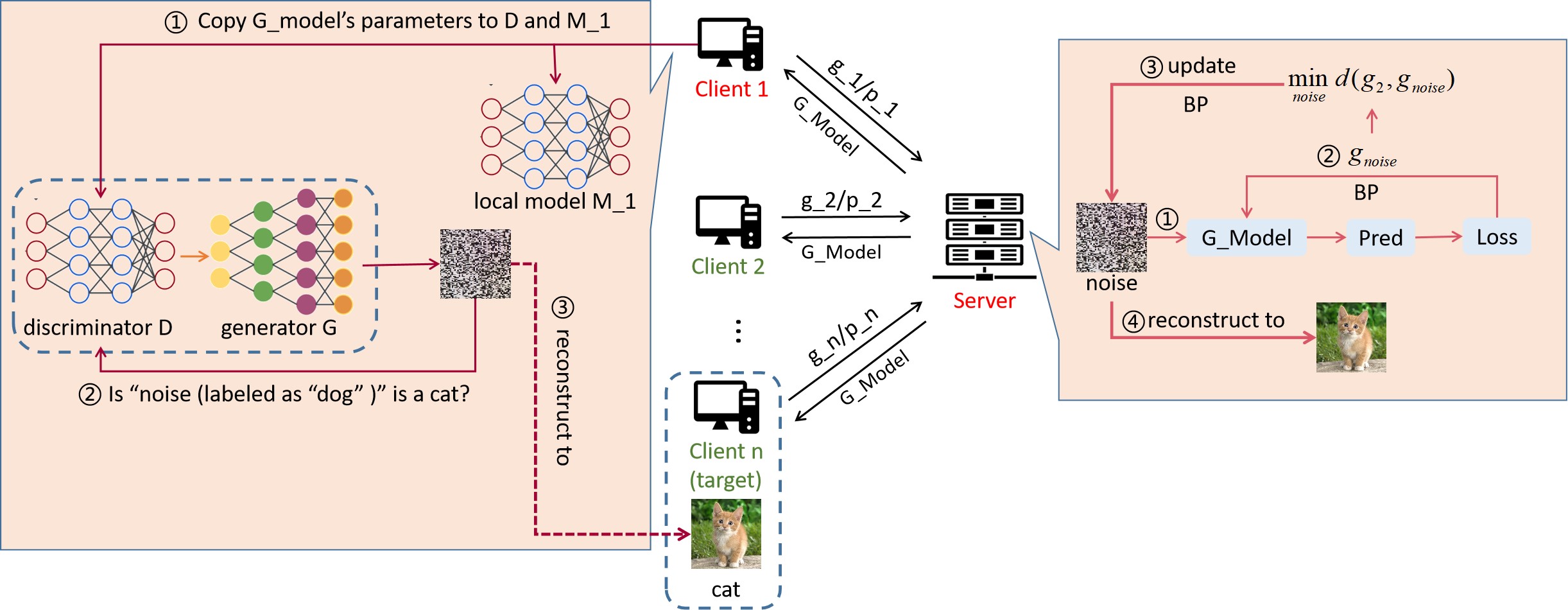}
	\caption{The paradigm of Gradients/Parameters-based RAs. Right: A general process of RAs using the gradient uploaded by the client when the server acts as the attacker; Left: Malicious users (attackers) use wrong labels to induce other participants (victims) to release information about target categories, and use GAN to capture this information to steal privacy.}
	\label{fig11}
\end{figure}

$1$) Gradients/Parameters-based attacks. In this attack, attackers will design a fitting scheme to fit the obtained gradients, and use it as an loss function to iteratively optimize a certain noise to reconstruct the victim's privacy. In addition, malicious servers can utilise the received local model parameters to reconstruct the client training set. This method is usually combined with GAN to get a satisfactory performance, as shown in Figure \ref{fig11}.

\begin{figure}[h]
	\centering
	\includegraphics[width=0.65\linewidth]{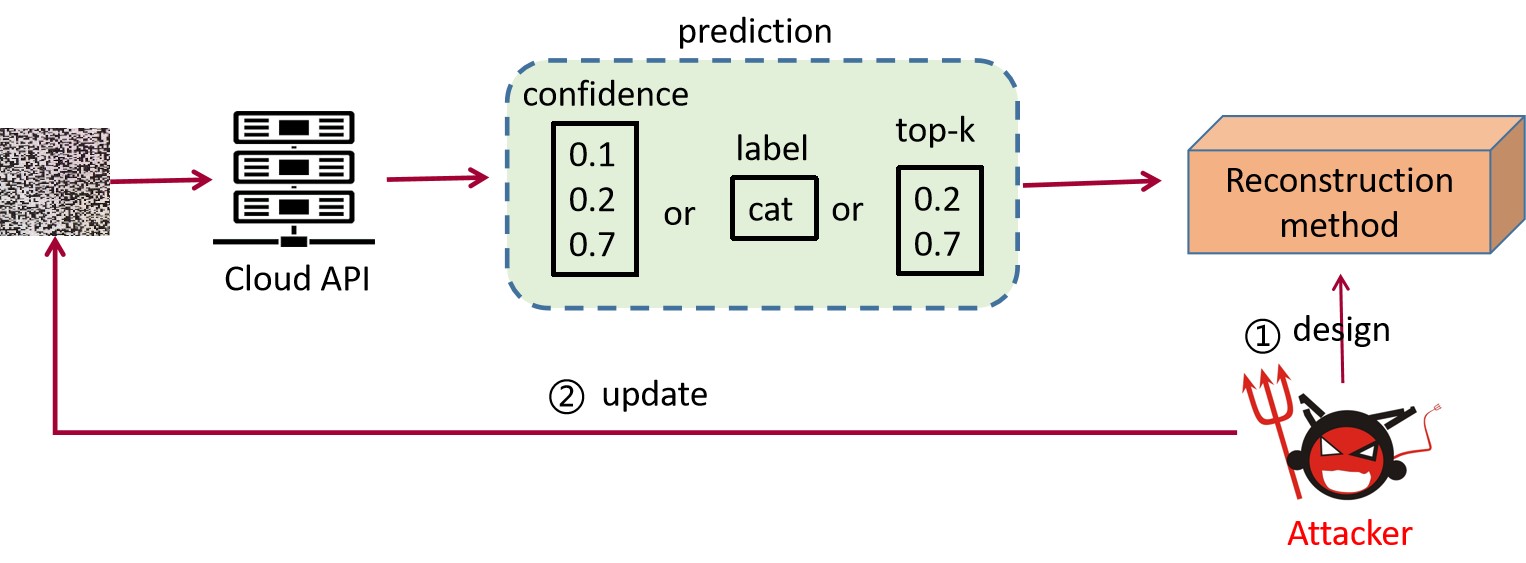}
	\caption{The paradigm of Model-based RAs. The attacker directly utilizes the model's outputs to carry out the attack. Generally, this cloud scenario is closer to an industrial scenario, and the target model is a black box for attackers.}
	\label{fig12}
\end{figure}

$2$) Model-based attacks: This type of attack usually uses the model's output (such as the prediction vectors or labels) to reconstruct a certain type of data feature of the target training set. For example, perform RAs against a published cloud model, as shown in Figure \ref{fig12}.

\textbf{Methodology of RAs}

In $2014$, RAs were first proposed in the context of genetic privacy by Fredrikson \emph{et al.} \cite{fredrikson2014privacy}, an attack capable of estimating someone's genotype through black-box access to the model. In $2015$, they applied RAs to a more complex dataset and evaluated using higher dimensional data (images) with white-box scenario \cite{fredrikson2015model}. Experiments show that the image can be reconstructed only using the target model's confidence vector output. However, this attack is only effective for simple machine learning models such as decision tree and two-layer neural networks. If combined with numerical approximation techniques, their proposed RA can perform attacks in a black-box setting, albeit with reduced attack efficiency.

Hidano \emph{et al.}\cite{hidano2017model} enhanced RA using poisoning attacks \cite{zhang2020online,biggio2012poisoning} and proposed general model inversion (GMI). Their method achieves comparable performance to that of \cite{fredrikson2015model} in black-box scenario. i.e., when inferring target data $X$, attackers can infer $X=\{x_1,x_2,...,x_n\}$ completely without knowing part of the information $X_{part}=\{x_e,x_f,...,x_j\}$ where $1 \leq e \leq j \leq n$. Specifically, for models that can be trained online (such as recommender system), attackers first transform the model into a target model that is more vulnerable to RAs through poisoning attacks. Then, reconstruct $X$ with auxiliary information and the target model's outputs. However, their research is still focused on traditional machine learning models. 

\begin{figure}[h]
	\centering
	\includegraphics[width=0.9\linewidth]{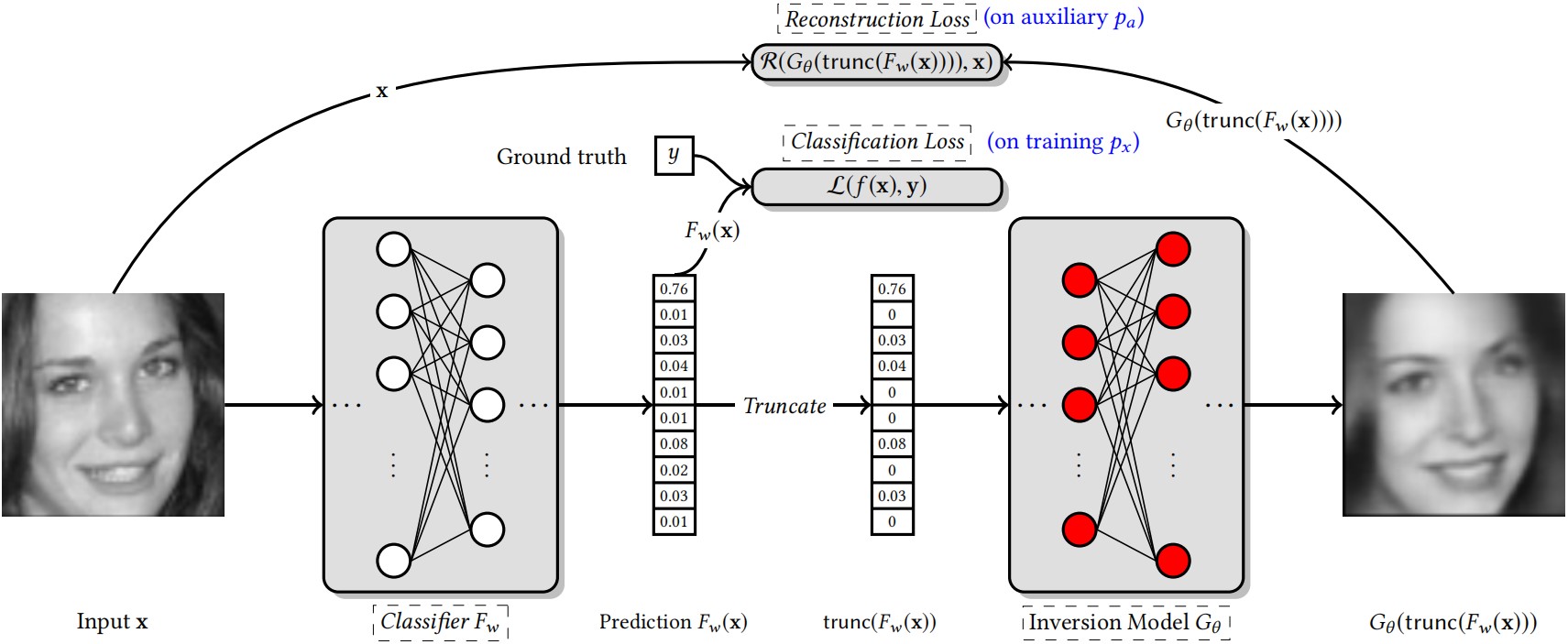}
	\caption{Attribute inference attack with deep neuron network.}
	\label{fig10}
\end{figure}

Yang \emph{et al.} \cite{yang2019neural} proposed a method based on truncated prediction vectors and optimizing the reconstruction loss. Attackers can implement effective RAs without knowing the target model's structure, parameters, and training set distribution. Specifically, they treat the prediction vectors of the target model as a feature of the original input that can help attacks to reconstruct the model input through a generative (reconstruction) model. As shown in Figure \ref{fig10}. They successfully extended RAs to high-dimensional data and deep neural networks.

Figure \ref{fig10} can be formulated as Formula \ref{eq4}. First, the prediction values $f(x)$ are truncated using a threshold method to retain representative feature values. Then, the inversion model $G$ is trained with a reconstruction loss $d(\cdot,\cdot)$ based on the truncated features. Finally, the optimized $G$ can reconstruct the target model's input through the target model's output.
\begin{equation}
	\min _\theta d(x, G(\operatorname{trunc}(f(x)) ; \theta))
	\label{eq4}
\end{equation}

However, cloud services in practical scenarios usually only return a predicted label to users (potential attackers) instead of a confidence vector, such as the Amazon Rekognition API \cite{API}. Compared with data reconstruction based on confidence vectors, label-based reconstruction is much more challenging, because the labels hardly contain any information about the target data. Therefore, the above methods are difficult to pose a threat to existing cloud services. To further explore the threat of RA, Ye \emph{et al.} \cite{ye2022label} proposed a label-only model inversion attack for the first time. Their attack is divided into $5$ steps as follows.

\begin{itemize}
	\item Step $1$. For a given target data record $\mathbf{x}^*$, the attacker collects a set of data records $D_{a u x}$ that belong to the same class as $\mathbf{x}^*$. The attacker then adds Gaussian noise to each record in $D_{a u x}$, and feeds the noisy set to the target model $T$ to test $T$'s classification error rate $\mu$. 
	\item Step $2$. The attacker uses $\mu$ and Gaussian Isoperimetric Inequality \cite{gilmer2019adversarial} to compute the distance $d$ from $\mathbf{x}^*$ to the decision boundary of model $T$. 
	\item Step $3$. The attacker adopts the dataset $D_{a u x}$ to train a shadow model $S$. 
	\item Step $4$. The attacker uses the weights of model $S$ and the distance $d$ to approximately compute the confidence score vector y of record $\mathbf{x}^*$. 
	\item Step $5$. Finally, the attacker uses $D_{a u x}$ to train an attack model $A$ to reconstruct the target record $\mathbf{x}^*$.
\end{itemize}
Even if the reconstruction effect is limited, they also proved that the label-only RA is feasible.

Undoubtedly, label-only reconstruction attacks are closer to reality. In addition, it is worth noting that the above RAs directly optimized the data in the high-dimensional space, which can easily make the optimization fall into the local minimal that is not corresponding to any meaningful image. Therefore, Kahla \emph{et al.} \cite{kahla2022label} designed a RA based on zero-order optimization. Their intuition is that "the more class-representative the data, the farther it will be from the decision boundary." For example, if the cat's decision boundary in a model is considered a circle, the cat on the particle is a "typical cat." However, it is difficult to find the optimal solution directly in high-dimensional space, thus, they proposed that a GAN can be trained using a public dataset to reconstruct a set of codes $Z$ in low-dimensional latent space to high-dimensional space to get $X^\prime$. Then utilize the target model to get the $X$s' label, and leverage these labels for the zero-order optimization to update $Z$. In the end, the representative data of the target category can be reconstructed.

So far, there have been many studies about reconstruction attacks, but RA in white-box scenarios is of little value. In contrast, the research value of RA in black-box scenarios is considerable, but the overall reconfiguration effect could be better. In addition, RAs in other fields are still in their infancy, such as speech recognition, natural language processing, etc.

\textbf{RAs in Federated Learning}

Compared with RAs in other scenarios, RAs in federated learning pose a more significant threat to privacy. For example, when the attacker is the server, s/he knows all the information about the target model, including parameters, structure, gradients, etc.; when the participants act as attackers, they can use poisoning attacks to change the training direction of the target model, making the model vulnerable. Zhu \emph{et al.} \cite{zhu2019deep} proposed a RA called Deep Leakage from Gradients (DLG) based on gradient fitting for federated learning scenarios, such as Formula \ref{eq5}. Specifically, they believe there is an injective relationship between the gradients and the input data, i.e., a set of gradients corresponds to a specific training data set. Thus, the attacker can inverse the target data precisely if a bunch of noises' gradients is fitted to the target gradients and updates these noises with the BP algorithm. This method can perform pixel-wise/token-wise reconstruction on the client's image/text training set. Moreover, their method can infer the label of the reconstructed data, even with low accuracy. In $2020$, Zhao \cite{zhao2020idlg} \emph{et al.} pointed out that if the last layer of the machine learning model uses a non-negative activation function, then the label corresponding to the input can be obtained directly according to the positive or negative gradient, which can be used to enhance the attack efficiency of DLG. However, These gradients fitting-based methods in FL are usually based on some strong assumptions. For example, the attacker needs to know the target data's batch size or the activation function type, which makes these attacks impractical.

\begin{equation}
	\mathbf{x}^{\prime *}, \mathbf{y}^{\prime *}=\underset{\mathbf{x}^{\prime}, \mathbf{y}^{\prime}}{\arg \min } d (\nabla W^{\prime},\nabla W)=\underset{\mathbf{x}^{\prime}, \mathbf{y}^{\prime}}{\arg \min }\left\|\frac{\partial \mathcal{L}\left(F\left(\mathbf{x}^{\prime}, W\right), \mathbf{y}^{\prime}\right)}{\partial W}-\nabla W\right\|^2,
	\label{eq5}
\end{equation}
where ${x}^{\prime}$ is the initial noise (such as Gaussian noise), and it is continuously optimized by fitting the gradient $\nabla W$, so that it can be reconstructed into ${x}^{\prime *}$ similar to the private data corresponding to $\nabla W$. In \cite{geiping2020inverting}, Geiping \emph{et al.} define the distance metric in Equation \ref{eq5} as cosine similarity, as in Equation \ref{eq6}. And they use the total variational loss as a regularization term to make the semantic of the reconstructed image clearer.
\begin{equation}
	\arg \min _{x^{\prime} \in[0,1]^n}\left\{1-\frac{\left\langle\nabla_\theta \mathcal{L}(x, y), \nabla_\theta \mathcal{L}\left(x^{\prime}, y\right)\right\rangle}{\left\|\nabla_\theta \mathcal{L}(x, y)\right\|_2 \cdot\left\|\nabla_\theta \mathcal{L}\left(x^{\prime}, y\right)\right\|_2}+\alpha \mathrm{TV}(x)\right\},
	\label{eq6}
\end{equation}
where $\langle\cdot\rangle$ is the inner product in $\mathbb{R}^n$ and $\|\cdot\|_2$ is the $L_2$-norm. $\alpha$ is a hyperparameter scaling the total variation penalty over the image. Usynin \emph{et al.} \cite{usynin2022beyond} added two regularization terms, style loss and feature loss, to gradient-based RAs based on the prior knowledge of the attacker. Their method outperforms existing gradient-based methods both qualitatively and quantitatively.

% Please add the following required packages to your document preamble:
% \usepackage{multirow}
\begin{table}[]
	\centering
	\tiny
	\begin{tabularx}{\linewidth}{|X<{\centering}|X<{\centering}|X<{\centering}|X<{\centering}|X<{\centering}|X<{\centering}|X<{\centering}|}
		\hline
		Year                  & Literature & Scenario & Target model type & Applied technology & Countermeasure-inclusive & Knowledge-level \\ \hline
		2015 &  Fredrikson \emph{et al.} \cite{fredrikson2015model}&Cloud Service&Decision tree; Simple neuron network& Maximum a posteriori estimator& Yes &Black/White-box \\  \hline
		
		\multirow{3}{*}{2017} & Hitaj \emph{et al.} \cite{hitaj2017deep}  & Federated learning &  CNN& GAN & Yes & White-box  \\ \cline{2-7} 
		& Hidano \emph{et al.} \cite{hidano2017model}&Online ML system & Traditional model & Poisoning attack  & No      & Black-box  \\ \hline
		
		\multirow{3}{*}{2019} & Yang \emph{et al.} \cite{yang2019neural}  & Cloud Service &  CNN& Truncated prediction; Reconstruct loss & No & Black-box  \\ \cline{2-7} 
		&  He \emph{et al.} \cite{he2019model} &  Collaborative learning & CNN     & Fit intermediate outputs; Total variation loss & Yes   & Black/White-box      \\ \cline{2-7} 
		&  Aivodji \emph{et al.} \cite{aivodji2019gamin} &  Cloud Service & CNN     & GAN & Yes   & Black/White-box      \\ \cline{2-7}  
		&  Zhu \emph{et al.} \cite{zhu2019deep}    &  Federated learning & CNN; RNN           & Fit gradients                 & Yes      &  White-box     \\ \hline
		
		\multirow{3}{*}{2020} &Wu \emph{et al.} \cite{wu2020evaluation}& Medical deep learning & CNN   & Fit intermediate outputs  & Yes & Black-box  \\ \cline{2-7} 
		&Zhao \emph{et al.} \cite{zhao2020idlg}& Federated learning  & CNN   & Fit intermediate outputs  & No & White-box  \\ \cline{2-7} 
		&Chen \emph{et al.} \cite{chen2020improved}& Cloud Service  & CNN   & GAN  & No & White-box  \\ \cline{2-7} 
		&Geiping \emph{et al.} \cite{geiping2020inverting}& Federated learning  & CNN   & Fit gradients; Cosine similarity; Total variation  & No & White-box  \\ \cline{2-7} 
		&Mo \emph{et al.} \cite{mo2020querying}&Cloud Service  & CNN   & Fit intermediate outputs  & No & Black-box  \\ \cline{2-7} 
		&Salem \emph{et al.} \cite{salem2020updates}&Online ML system  & CNN   & VAE; Model posterior difference  & Yes & Black-box  \\ \cline{2-7} 
		&Zhang \emph{et al.} \cite{zhang2020secret}& Cloud Service & CNN   & Fit intermediate outputs; Fit target model's outputs  & No & Black-box  \\ \hline

		\multirow{3}{*}{2021} &Zhao \emph{et al.} \cite{zhao2021exploiting}& Explainable AI in cloud service & CNN   & Reconstructed explanation loss  & No & Black-box  \\ \cline{2-7} 
		&Hernandez \emph{et al.} \cite{hernandez2021prid}& Hyperdimensional computing system  & CNN; AdaBoost \cite{wang2021improved}   & Decoder; Feature replacement  & Yes & White-box  \\ \cline{2-7} 
		&Khosravy \emph{et al.} \cite{khosravy2021model}& Facial recognition system  & CNN   & Fit target model outputs  & No & Black/White-box  \\ \cline{2-7}  
		&Carlini \emph{et al.} \cite{carlini2021extracting}& Cloud service  & GPT-2   &  MIA; Internet search  & No & Black-box  \\ \cline{2-7} 
		&Yin \emph{et al.} \cite{yin2021see}& Federated learning & CNN   &  Fit gradients; Fidelity and group consistency regularization & No & White-box  \\ \cline{2-7} 
		&He \emph{et al.} \cite{he2021stealing}& Cloud service & GNN   & Entropy; Node similarity; ST & No & Black/White-box  \\ \cline{2-7} 
		&Wu \emph{et al.} \cite{wu2022linkteller}& Cloud service & GNN   & Nodes' influence value; Sort & No & Black-box  \\ \cline{2-7} 
		&He \emph{et al.} \cite{he2021stealing}& Cloud service & GNN   & Entropy; Node similarity; ST & No & Black/White-box  \\ \cline{2-7} 
		&Wang \emph{et al.} \cite{wang2021variational}& Cloud service  & CNN   & Variational inference; GAN  & No & White-box  \\ \hline

		\multirow{3}{*}{2022} &Hatamizadeh \emph{et al.} \cite{hatamizadeh2022gradvit}& Facial recognition system  & Vision transformers   & Fit gradients; Image prior; Patch prior  & No & White-box  \\ \cline{2-7} 
		&Usynin \emph{et al.} \cite{usynin2022beyond}& Collaborative learning & CNN & Fit gradients; Fit features and style of target data  & No & White-box  \\ \cline{2-7} 
		&Khosravy \emph{et al.} \cite{khosravy2022model}& Cloud service & CNN & $\alpha$-GAN; Poincare loss   & No & White-box  \\ \cline{2-7} 
		&Olatunji \emph{et al.} \cite{olatunji2022private}& Cloud service & GNN & Feature explanation; Pairwise similarity   & Yes & Black/White-box  \\ \cline{2-7} 
		&Ye \emph{et al.} \cite{ye2022model}& Transfer learning & CNN & Shadow model; Generator  & No & Black-box  \\ \cline{2-7} 
		&Kahla \emph{et al.} \cite{kahla2022label}& Cloud service & CNN &  Boundary repulsion; GAN & No & Black-box  \\ \cline{2-7} 
		&Ye \emph{et al.} \cite{ye2022label}& Cloud service  & CNN  & Shadow training; CDB & Yes& Black-box  \\ \hline
	\end{tabularx}
	\begin{tablenotes} 
		\item \textbf{CDB}: Calculate the distance from the target sample to the decision boundary; \textbf{ST}: Shadow training; 
	\end{tablenotes}
\end{table}

\textbf{RAs in Other Models}

In $2021$, He \emph{et al.} \cite{he2021stealing} first proposed a reconstruction attack against graph models, inferring whether there is a link between two given node pairs in the target dataset. They assume that the attacker has some prior knowledge, thus the ST-based reconstruction attack can be constructed simply by measuring the similarity between nodes. Wu \emph{et al.} \cite{wu2022linkteller} calculate the influence value between pairs of nodes, and then sort the nodes according to the influence value. The two nodes with the highest influence values are linked. GraphMI \cite{zhang2021graphmi} aims at reconstructing the adjacency matrix of the target graph given white-box access to the trained model, node features and labels. Olatunji \emph{et al.} \cite{olatunji2022private} show that additional knowledge of post-hoc feature explanations substantially increases the success rate of these attacks

So far, reconstruction attacks have been proven to pose threats to many fields, the most typical of which is the cloud service scenario. However, RAs mainly attack image-related machine learning models, such as image classifiers and generators. Thus, RSs' main attack objects are CNN and its extended versions. Researchers currently have relatively few RA studies on text/speech-related models \cite{carlini2021extracting}. 

\subsection{Model Extraction Attacks}
\begin{figure}[h]
	\centering
	\includegraphics[width=0.9\linewidth]{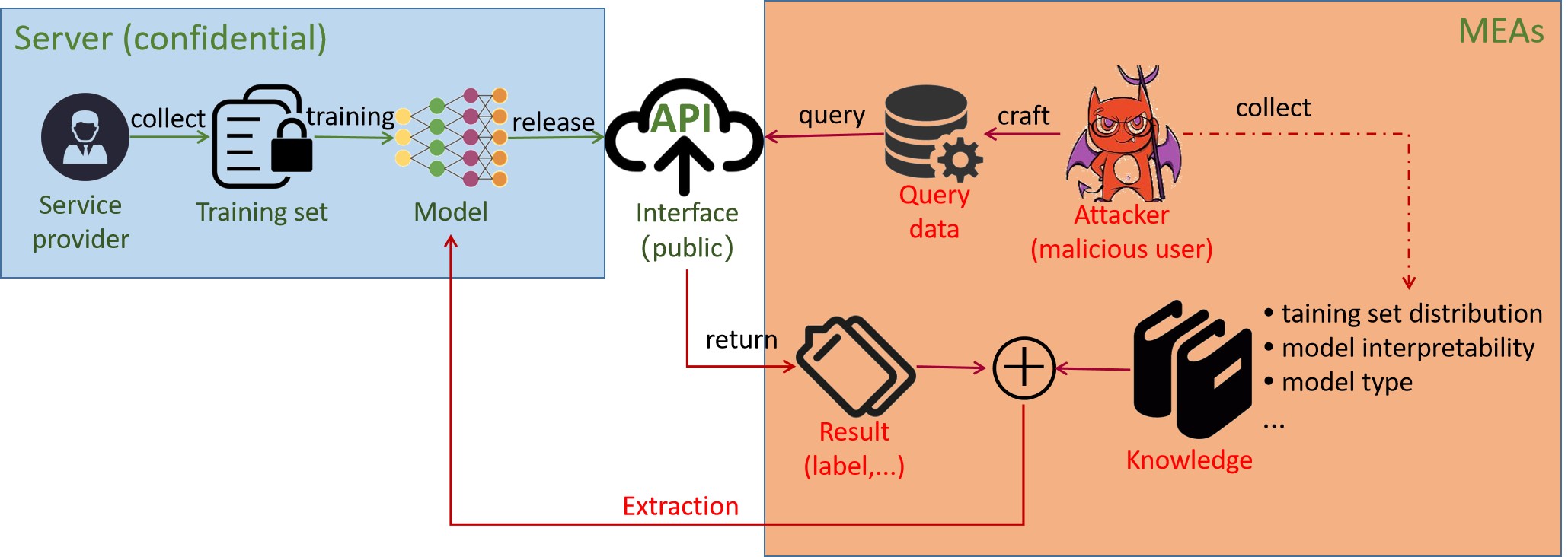}
	\caption{Diagram of model extraction attacks.}
	\label{fig13}
\end{figure}

 Unlike the above-mentioned inference attacks, MEAs are not meant to steal data privacy. Conversely, MEAs aim to steal the privacy of trained models, such as parameters, structure, capabilities, etc. A common scenario is that attackers are usually only allowed to access the target model through an open API to get its output. Some works assume that an attacker may know other knowledge about the model, such as the model's type, interpretability, etc., as shown in Figure \ref{fig13}. With the popularity of MLaas, many machine learning models with practical commercial value are considered expensive, confidential, and valuable private intellectual property rights, because the model's owner may spend extremely high costs collecting training sets and debugging the model. Attackers can replicate the ability of these machine learning models through MEAs, thereby infringing the rights of the model owners. In addition, MEA can help attackers implement other attacks, such as using the extracted model to craft adversarial samples to implement adversarial attacks. Due to the transferability of adversarial samples, the adversarial samples obtained using the extracted model can also be used to attack the target model, and generally, the attack is more difficult to be detected.

Generally speaking, there are two ways to extract the target model, namely:

$1)$ Extract model parameters. "Parameters" represent the "weight parameters" of each node of the model and the "hyperparameters (such as learning rate)" during model training. This attack method usually assumes that the attacker knows the structure of the model (or at least the model type). In fact, as long as the attacker knows the model parameters, the functionality of the model is also stolen.

$2)$ Extract model structure. After knowing the model structure, the attacker can carry out more threatening attacks.

$3)$ Extract model functionality. Unlike the previous two ways, this way focuses on imitating the model's behavior, such as classification, generation, regression, etc., rather than the model's attributes. For example, the target model is an SVM model for classification, but an attacker can use a deep learning model to simulate the capabilities of this SVM.

We now elaborate on the above three ways.

\textbf{$1)$ Extract the weight parameters and hyperparameters of target model:}  Tramer \emph{et al.} \cite{tramer2016stealing} first proposed the concept of model extraction. In their proposed MEAs, attackers perform massive queries against cloud models using carefully crafted or publicly available data. Then use the obtained output and query data to construct input-output pairs to obtain a series of parametric equations, which can be understood as multivariate equations. By solving these equations (e.g., least squares), attackers can infer the internal parameters of the target model. However, the MEA they proposed requires the attacker to know the output (confidence, label) and type (SVM, neural network) of the target model to implement the attack successfully. Furthermore, due to the inherent computational complexity of methods based on solving multivariate equations, the first proposed MEA is only suitable for simple models with few parameters and is ineffective for DNN models. Nowadays, deep learning has been integrated into every aspect of our lives. As a commercial asset, the use of deep models has gained popularity. For a deep neuron network, the amount of its parameters is tremendous, and the learned mapping relationship is also highly complex. Hence, it is not feasible to steal the model only by solving multivariate equations.

Apart from model parameters, hyperparameters are also extremely worthy of attention, because different hyperparameters often make the same model have different performances. The selection of optimal hyperparameters is likely to cost a lot to the model trainer, thus they are as confidential as the model parameters. In $2019$, Wang \emph{et al.} \cite{wang2018stealing} first proposed an attack to steal hyperparameters in the objective/loss function. For machine learning algorithms using such hyperparameters, the model parameters are usually learned from the local minimum of the objective/loss function. Therefore, the gradients of the objective function at the learned model parameters should be $0$, which encodes the relationship between the learned model parameters and hyperparameters, resulting in a linear overdetermined system (the number of equations is greater than the number of unknown variables). Finally, an attacker can use the linear least squares method to solve the overdetermined problem in the linear system. Similarly, their attacks apply only to simple machine learning algorithms, such as ridge regression, logical regression, support vector machines, and simple neural networks.

\begin{figure}[h]
	\centering
	\includegraphics[width=0.5\linewidth]{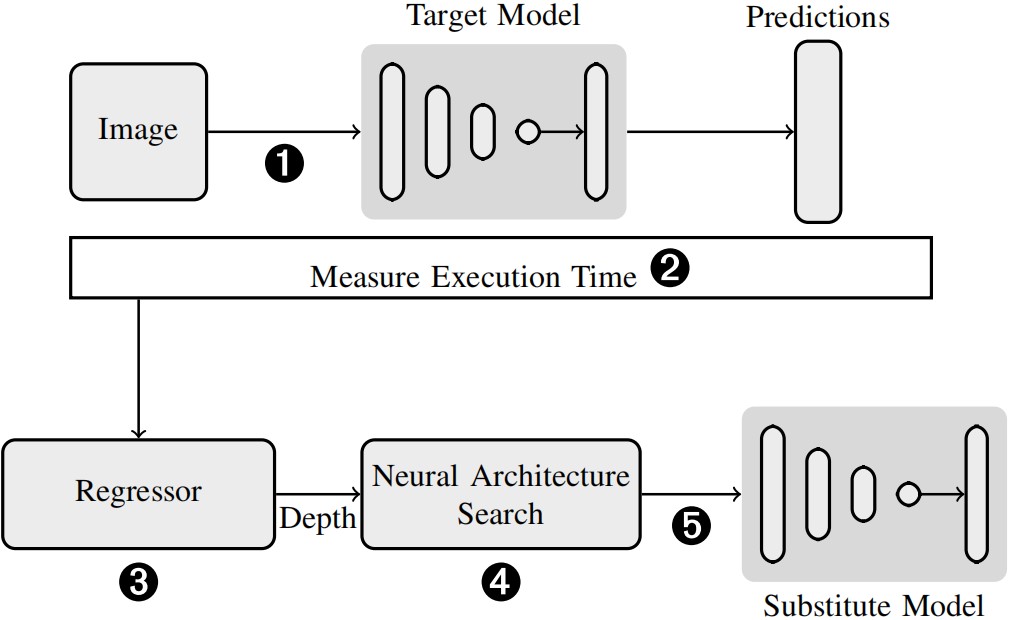}
	\caption{Diagram of model extraction attacks.}
	\label{fig14}
\end{figure}

\textbf{2) Extract model structure:}To infer the architecture of cloud models, Duddu \emph{et al.} \cite{duddu2018stealing} proposed a timing side channels-based model architecture search method named StealNN. Constructing an architecture equivalent to the target model is a complex search problem. However, the authors presented how reinforcement learning effectively reduces the search space and reconstructs optimal substitute architectures close to the target model. StealNN can be divided into five steps, as shown in Figure \ref{fig14}, where $(1)$ the attacker queries the target model and obtains the output of the model; $(2)$ the attacker measures the execution time of the target model; $(3)$ takes the execution time as the Regressor's input to predict the depth of the target model; $(4)$ Search the structure of the target model, and use the obtained depth to reduce the search space; $(5)$ Use reinforcement learning to generate the optimal structure in the search space as a substitute model of the target model. The disadvantage of stealNN is that it requires the attacker to know what hardware the target model is trained on, and the attacker needs to use the same hardware to build the timing profiles. 

Recently, Juuti \emph{et al.} \cite{juuti2019prada} proposed an iterative model extraction strategy called PRADA, which consists of many repeated rounds. Each round consists of three steps: $(1)$ querying the cloud-based model; $(2)$ training a substitute model using the obtained input-output pairs; $(3)$ synthesizing new query data based on Jacobian-based data augmentation (JBDA) using the substitute model.  Different from other methods, PRADA is validated by the transferability of adversarial examples.

\textbf{3)Extract model functionality:} When attackers do not have any knowledge about the target model and the target training set, they can only focus on stealing the functionality of the model rather than the model parameters or structure. Shi \emph{et al.} mentioned in \cite{shi2017steal} that it is feasible to fit the functionality of simple models with more complex machine learning models, such as fitting Bayesian classifiers with deep neural networks (DNNs). However, the reverse process is not feasible, such as fitting a DNN with a Bayesian classifier. They evaluated this idea on a text classification task. To demonstrate that MEAs can also extract deep neural networks, Orekondy \emph{et al.} proposed Knockoff Nets \cite{orekondy2019knockoff}. Like the previous method, Knockoff Nets also need to query the API to obtain input-output pairs and train a substitute model. The difference is that they demonstrate that appropriate complex structures can reproduce the functionality of DNNs. For example, VGG-19 can be used to extract ResNet-10. In addition, the attacker also needs a vast natural dataset as the query dataset, and datasets such as ILSVRC can be considered. Orekondy \emph{et al.} also filter query samples to achieve high fidelity using reinforcement learning.

\begin{figure}[h]
	\centering
	\includegraphics[width=1\linewidth]{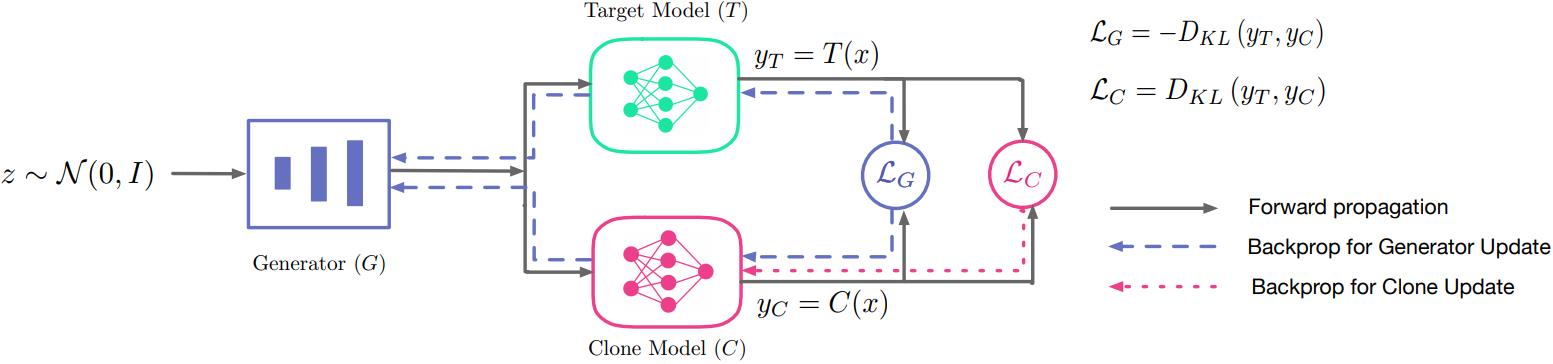}
	\caption{MAZE attack setup.}
	\label{fig15}
\end{figure}

In $2021$, Kariyappa \emph{et al.} \cite{kariyappa2021maze} proposed a Data-Free MEA called MAZE. Unlike Orekondy \emph{et al.}, they used GAN to allow attackers to implement MEA without a large amount of natural data. Specifically, MAZE first constructs an initial substitute model $C$. Then use a generative model $G$ to generate query data to query the target model $T$ and the substitute model $C$, respectively, and obtain labels $T(x)$ and $C(x)$. Second, they used the idea of game theory and set the objective function of $C$ to minimize the distance between $T(x)$ and $C(x)$ to imitate the behaviour of $T$. For $G$, it needs to generate good enough samples to fool $C$. From the perspective of MEAs, better samples can help $C$ to better imitate the functionality of $T$. MAZE is shown in Figure \ref{fig15}.

The above three model extraction methods show that most MEAs rely on the input-output pair consisting of query data and the output of the target model to carry out attacks. Therefore, how to construct efficient query data is a valuable problem. At present, there are four \textbf{main ways to build a query dataset.}

\textbf{1) Random noise:} For example, random sampling from a normal distribution, or sampling from a distribution set composed of multiple distributions, such as sampling from Gaussian distribution, Poisson distribution, and uniform distribution simultaneously. The input-output pair obtained by this method is only valid for simple machine learning models (such as traditional machine learning models and simple neural networks).

\textbf{2) Random natural data.} In this case, it is assumed that the attacker only has access to publicly available large-scale datasets, i.e., random natural images or non-problem domain images available on the Internet. Natural images have more complex distributions and diverse features, and therefore a better choice than "random noise".

\textbf{3) Synthetic data.} In this case, the attacker usually has some knowledge of the target training set, such as its marginal distribution, based on which the attacker can artificially synthesize query data. However, some studies have shown that even if attackers do not know this knowledge, they can use some means to make the target model leak relevant information, such as using model inversion attacks.

\textbf{4) A dataset with the same/approximate distribution as the target training set.} In this case, it is assumed that the attacker has data from the same/approximate distribution as the target training set, i.e., the attacker has knowledge of the distribution of the target training set. Undoubtedly, input-output pairs composed of such data can better capture the functionality of the target model that maps the source domain to the target domain.

\textbf{MEAs in Other Applications}

Like the limitations of several other inference attacks, the research of MEAs mainly focuses on classification models. However, there are also a small number of studies that have broken through this limitation. For example, in \cite{milli2019model}, except for constructing input-output pairs, it explores the risk of the API providing users with the input's gradients as the interpretability of its output. The purpose of providing interpretability is to allow users to trust the API's output. But unfortunately, this will lead to the vulnerability of the cloud model. Chandrasekaran \emph{et.al} \cite{chandrasekaran2020exploring} found that active learning can help researchers better understand MEAs, thereby discovering stronger MEAs or effective defense methods, because active learning has certain similarities with MEA. krishna \emph{et.al} \cite{krishna2019thieves} studied how to extract BERT-based natural language models. Unlike image task models, queries to BERT-based APIs do not require natural or semantically informative data. Random sequence queries combined with task-specific heuristics are sufficient to extract natural language models efficiently.

It is worth noting that model extraction attacks on generative models have not been studied. Although some researchers believe it is possible to use the generated data to directly construct a training for training surrogate models, no research has confirmed this.

\section{Defence Strategy in MLaaS}
Since the target model and target training set are essential assets of cloud service providers in MLaaS, to maintain a benign business environment and prevent privacy leaks, it is indispensable to defend against the above inference attacks. This section presents these countermeasures. However, due to limited space, this section mainly expounds the idea of these countermeasures rather than their specific process.
\subsection{Genaral Defenses}
Although above-mentioned inference attacks are different, they all have similarities. Therefore, some defense methods can be used to defend against various inference attacks, and this subsection discusses some such general methods.

\textbf{Perturbation} 

Existing studies have shown that a vital factor for the success of inference attacks is that attackers obtain the (partial) mapping pattern of the target model to map the source domain (training set) to the target domain (label). Therefore, a straightforward coping strategy is to break this mapping pattern. And because most inference attacks rely on obtaining the output of the target model to implement, thus, the most direct method is to return to the users the final result of target model (i.e., hard label), Top-k confidence, or noised confidence \cite{lee2019defending,orekondy2019prediction,zheng2019bdpl}. Nevertheless, as discussed in the previous section, even hard labels can successfully make inference attacks \cite{juuti2019prada,orekondy2019knockoff}. 

In addition, differential privacy (DP) is one of the most widely used and effective perturbation method for privacy protection, proposed by Dwork \emph{et al.} \cite{dwork2014algorithmic} in $2006$. In the machine learning domain, some researchers refer to it as differential machine learning. It is defined as: a random function $M$ and a well-defined probability density $P$ satisfy $\epsilon-$differential privacy, if for all adjacent databases $D$ and $D^\prime$ and for any $S \in range(M)$, such as Equation \ref{eq12}.
\begin{equation}
	\left|\log \frac{P(M(D)=S)}{P\left(M\left(D^{\prime}\right)=S\right)}\right| \leq \epsilon,
	\label{eq12}
\end{equation}
There are currently \textbf{two main categories of DP-related privacy protection methods} used to counter inference attacks:

\textbf{$1)$ Private Aggregation of Teaching Ensembles (PATE): }
PATE is a privacy protection method based on the idea of ensemble learning \cite{papernot2016semi,hamm2016learning,abadi2017protection}. It does not use private data to train the cloud model directly, but uses the idea of retraining and noise addition to training the cloud model. The specific steps are mainly divided into three steps:
\begin{itemize}
	\item Step $1$. Dataset segmentation: divide the privacy dataset into $n$ parts, and use each to train $n$ target task models.
	\item Step $2$. (Customizable privacy step) Develop privacy mechanism: Most of the different PATE methods are optimized and changed in this step. Here we mainly present two typical methods \cite{pathak2010multiparty, hamm2016learning}: $1.$ Collect a public dataset $D^\prime$ similar in distribution to the private dataset, and use the voting mechanism in ensemble learning to ensemble the obtained $n$ models. Second, use the ensemble model to label the data in $D^\prime$. Next, train model $M$ with $D^\prime$. Finally, add differential privacy perturbation to the parameter $w$ of model $M$ to obtain model $M^\prime$, and publish $M^\prime$ as API. $2.$ Aggregate the obtained $n$ models according to Equation \ref{eq11} and release them as an API, where $\eta$ represents the differential perturbation of the Laplace distribution.
	\item Step $3$. In order to further reduce the risk of privacy leakage, service providers can also adopt the strategy of outputting Top-K confidence.
\end{itemize}

\begin{equation}
	\hat{\mathbf{w}}=\frac{1}{n} \sum_{i=1}^{n} \hat{\mathbf{w}}_j+\boldsymbol{\eta}
	\label{eq11}
\end{equation}

From the perspective of data privacy leakage, PATE significantly reduces the possibility of privacy being inferred. Because the public model is not directly trained with private data sets, it can have a defensive effect on inference attacks such as MIA and PIA. However, its shortcomings are also obvious: $(1)$ The segmentation of the data set means that the service provider must hold enough data, which is a challenge for small sample learning; $(2)$ Performing model aggregation, retraining, and adding perturbations will bring more computational overhead. Some studies directly add differential noise to the activation function or outputs of the model output layer, but the counter effect is not satisfactory \cite{lee2019defending}. 

\textbf{$2)$ Gradients Perturbation: }

Machine learning models are computed step-by-step from gradients obtained through the training set. Therefore, gradient processing may be an entry point if we want to preserve data privacy. Typically, most models are trained iteratively using the stochastic gradient descent (SGD) algorithm. In each iteration, a small amount of training data is extracted from the training set to calculate the model error. Then the BP algorithm is used to obtain gradients to update the model. The DP-SGD proposed by Abadi \emph{et al.} \cite{abadi2016deep} has made two modifications to the model mentioned above update process to allow the model to obtain differential privacy: $(1)$ Calculate the gradients for each training data separately (i.e., batch-size=$1$), and clip the gradients that are smaller than a specified threshold. $(2)$ Add Gaussian noise to the clipped gradients. While gradient clipping is common in deep learning and is often used as a form of regularization, it differs from DP-SGD as follows: clipping is performed on the average gradient over mini-batches, rather than clipping individual examples before averaging gradients.

The above two perturbation strategies reduce the risk of privacy leakage to a certain extent. Still, it cannot be ignored that the introduction of perturbation will reduce the model's accuracy. Therefore, achieving a satisfactory trade-off before model utility and privacy is one of the main concerns of researchers at present.

\textbf{Encryption} 

As one of the most classic privacy protection methods, encryption has been widely used in various fields. Fully homomorphic encryption (FHE) \cite{viand2021sok} is the most promising encryption method that can be applied to machine learning. FHE can make the calculation of ciphertext completely homomorphic with plaintext. Thus, a trusted third party is not necessary when using it for privacy protection, which makes data very safe during transmission and processing. However, FHE will increase the computing and storage overhead geometrically, hence it is challenging to use it in combination with deep learning. Hence there are currently many FHEs that have improved efficiency, such as  HElib, FHEW, and HEEAN \cite{halevi2020design,ducas2015fhew}, but it is still difficult to apply them to MLaaS scenarios. Other encryption methods, such as Secure Multi-party Computation (SMC), are usually used to solve the problem of multi-party collaborative computing, such as multi-party collaborative training of a machine learning model in MLaaS. However, the mainstream business of MLaaS is mainly oriented to data prediction and processing, while SMC does not fit these service scenarios. If readers want to know more about the encryption method, please refer to \cite{sun2018private,lee2022privacy,wood2020homomorphic}.

\textbf{Adversarial Defences}

The idea of adversarial defense is to add the attack method as a penalty item to the loss function of model training. In theory, the adversarial defense can be used in almost all privacy security scenarios. For example, in the \cite{nasr2018machine}, Nasr \emph{et al.} regularized the target model based on the min-max game during the model training process, so that the model's output for the target data is indistinguishable from other data under the same distribution, thereby resisting membership inference attacks. This technique not only protects privacy, but also increases the model's generalization. Theoretically, many privacy attacks could be used as adversaries to improve against during the training of the target model. Nevertheless, we must be cautious about using state-of-the-art attacks as penalty items. Because some studies have shown that state-of-the-art attacks may increase the model's vulnerability, and even its privacy security is not as good as the undefended model \cite{madry2017towards, song2019privacy}.

\textbf{Observing Input}  

In addition to the above three general methods, cloud service providers can also detect potential attacks by detecting query data. This defensive strategy does not compromise the utility of the target model and is usually more efficient. Since many inference attacks rely on many queries, it is a very effective way to limit the number of user queries. In addition, the cloud service provider can also detect the distribution of query data (based on the distribution of the training set). If there are many query requests with abnormal distribution for the same user or at the same time, it can be judged that the cloud service is under attack. The Kesarwani \emph{et al.} \cite{kesarwani2018model} found that query summaries can be built for each client to compute the feature space covered by client queries. Similarly, PRADA proposed in \cite{juuti2019prada} computes the minimum distance between query data and all previous samples in the same class to rule out malicious queries.

\textbf{Machine Unlearning}  

The "General Data Protection Regulation (GDPR)" promulgated by the European Union gives citizens the right "to request that their data be deleted from the database". However, what if this data was used to train a machine learning model? So the next natural question is: what if this data must also be "forgotten" from the AI model? The easiest way is to remove the target data from the training set and retrain the model. However, this process consumes much computational cost, making it unfeasible in practical scenarios. Therefore, how to delete specific private information from the target model by technical means is an important research direction.

Machine Unlearning was first proposed by Cao \emph{et al.} \cite{cao2015towards} They converted the machine learning algorithm into an additive form to effectively forget about the data trace. The method can also resist data pollution attacks.\cite{baumhauer2022machine} proposed to forget logit-based classifiers via a linear transformation of the output logits. Nevertheless, this method leaves traces of data samples in the weights of the neural network model. \cite{guo2019certified} used the idea of differential privacy to remove data and provides an algorithm for convex problems based on second-order Newton updates.

\subsection{Targeted Defenses}
Some defense methods cannot defend against multiple inference attacks simultaneously, but they are effective against a specific attack. In this section, we will present readers with several targeted defense methods.

%数字水印  MEA
\textbf{Watermarking Techniques for MEAs:} Leveraging the inherent generalization and memory capabilities of neural network architectures, watermarking techniques allow models to embed carefully crafted watermarks during the training and prediction stages to verify the ownership of the target model, thereby defending against model extraction attacks. Nagai \emph{et al.} \cite{nagai2018digital} introduced digital watermarking technology into the deep neural network training process for the first time, and proposed a digital watermarking technology for deep neural network ownership authorization. They used a parameter regularizer to embed watermarks in the model parameters, showing that the method can incorporate watermarks only by fine-tuning. Thus the performance of the network where the watermark is placed will not be compromised. Even after the model is fine-tuned or parameter pruned, the embedded watermark does not disappear. e.g., $65\%$ of the parameters are pruned, the watermark remains intact.

\textbf{Adversarial perturbation for MIAs:} The attack intuition of MIAs is that the behavior of the target model for training data and non-training data is different, and the attacker can use this behavior difference to judge whether the target sample belongs to the target training set. Interestingly, many studies on adversarial attacks show that the adversarial perturbation required to convert training data into an adversarial example is significantly more significant than non-training data. Therefore, some researchers proposed combining these two ideas to detect MIAs using adversarial perturbation. Specifically, in the first stage, MemGuard \cite{jia2019memguard} carefully designs a noise vector that can convert the confidence of the target model output into an adversarial example, which is likely to mislead the attacker's member classifier. In the second stage, MemGuard adds the noise vector to the confidence score vector with a certain probability, so that the confidence of the model output satisfies the given utility loss budget. This defense idea can also be used to against RAs.
\section{Challenges and Opportunities}
Although there have been many studies on inference attacks, with the development of machine learning, new problems have emerged. In this section, we will raise some promising research directions for readers from both attack and defense aspects.
\subsection{Attack}
\textbf{(1) Construction of query data sets:} Most of the existing inference attacks require a large number of queries on the target model. However, these query data usually come from public natural datasets, normal distribution, marginal distribution, etc. Thus, the distribution of query data is not consistent with the target training set, which gives the defender many ways to detect the attack. Therefore, how to effectively construct query data sets to implement attacks more effectively is a direction worthy of research.

\textbf{(2) Inference attacks in different machine learning scenarios:} Inference attacks have not been widely explored in various machine learning scenarios. For example, model extraction attacks for generative models, parameter extraction for reinforcement learning, inference attacks for ensemble learning, semantic segmentation models, etc. Current inference attacks almost all focus on classification models and single scenarios. However, MLaaS has many classic scenarios, such as collaborative learning, federated learning, segmentation learning, etc. In addition, the target model also has different categories such as generation, classification, regression, unsupervised, and semi-supervised. Therefore, we can consider expanding the research scope of inference attacks to understand potential privacy issues better.

\textbf{(3) Concealment of attack:} Inference attacks usually require a large number of queries on the target model, and these abnormal queries are easily detected by the server, so as to make a defensive response. In addition, the distribution of the query data is usually different from the distribution of the target training set, which is also easily detected and makes the query invalid.Some existing research aims to use a few queries/data-free to improve the concealment of attack, but there are still cases where the inference effect is not ideal. Therefore, improving the effect of inference attacks while reducing the number of queries is worth studying.

\textbf{(4) Model extraction in federated learning:} Since the global model in federated learning is shared by each client and server, there is no threat of model extraction attacks on the surface. However, each client in federated learning is independent of the other and does not communicate with each other. Therefore, we need to consider whether some methods or scenarios allow malicious clients to steal other clients' local models, and further obtain the victim's local private data. This might be a point worth noting.

\subsection{Defense}
\textbf{(1) Privacy-utility trade-off:} Most of the existing defenses will affect the performance of the target model, such as perturbation will reduce the model's accuracy, and encryption will increase the computational cost. Beyond that, defenses such as machine unlearning and adversarial machine learning have not gained widespread acceptance because researchers have found that they can lead to privacy breaches. These are points that can be studied and improved in the future.

\textbf{(2) Query detection:} In all inference attacks, the attacker needs to query the model to indirectly obtain the knowledge of the target model and the training set. These query data usually mimic the query distribution of benign clients, or use samples with low information gain and limited coverage of the feature space. How to effectively detect these more intelligent inference attacks needs further research.

\textbf{(3) Side-channel attack detection:} Side-channel attack strategies can be applied to model extraction attacks (such as timing side-channel and electromagnetic side-channel). Both can achieve a high attack success rate. However, no defense strategies are currently designed against such model extraction attacks. This should be a significant development direction.

\textbf{(4) Adversarial defense:} Although studies have shown that adversarial defense may strengthen privacy leakage, it does not affect its potential privacy protection scheme. For example, adversarial training is highly effective in adversarial attacks. It is worth noting that the failure of adversarial defenses may be due to the addition of wrong or inappropriate distance constraints, such as $l_p$, which are used for adversarial model training. Although $l_p$ perturbation constraints have been widely used in adversarial machine learning, there are still some limitations. 

\textbf{(5) Machine unlearning:} Machine unlearning has been a hot topic of privacy protection in recent years. Nevertheless, because existing research does not fully understand the black-box mode of machine learning, it is challenging to design a recognized and provable machine unlearning method. However, its potential for privacy protection cannot be ignored, and it is a direction worthy of research. In a sense, if the development of machine unlearning takes a leap, we can achieve personalized privacy in almost all machine learning scenarios.

\textbf{(6) Combination of machine unlearning and adversarial machine learning:} As two independent defense methods, machine unlearning and adversarial machine learning have been proven to have certain privacy protection effects. A research direction that can be referred to is to combine them, that is, to use adversarial machine learning to realize machine unlearning. Specifically, when we need to make the model forget specific training data, we can introduce an inference attack similar to MIA as a regular term to retrain the model. During training, the regularization term can help the model quickly forget the knowledge of the target data. This can solve the problem that the existing unlearning methods cannot forget the target data in a fine-grained manner.

\section{Conclusion}
This survey presents five major inference attacks in machine learning: MIAs, PIAs, RAs, AIAs, and MEAs. By reviewing a large amount of literature, we found that the existing taxonomy of inferred attacks is confusing, which will be unfavorable for subsequent researchers to study. Therefore, we first proposed a taxonomy called 3MP, which gave each inference attack a unique definition, thus clarifying the differences and connections of various inference attacks. Next, we overview and review the general methods and developments of various inference attacks from different MLaaS scenarios, model types, etc. Based on our observations, "inference attacks" are evolving towards stealthiness, making them harder to detect. In addition, the defender is relatively passive in this privacy game, that is, the attack method appears before the counter strategy. For those common defense strategies, the trade-off of privacy utility is still an open problem. In the end, we discussed some promising research directions in the future from the perspective of defense and attack, respectively.

%%k
%% The acknowledgments section is defined using the "acks" environment
%% (and NOT an unnumbered section). This ensures the proper
%% identification of the section in the article metadata, and the
%% consistent spelling of the heading. 3MAP
%\begin{acks}
%To Robert, for the bagels and explaining CMYK and color spaces.
%\end{acks}

%%
%% The next two lines define the bibliography style to be used, and
%% the bibliography file.
\bibliographystyle{ACM-Reference-Format}
\bibliography{reference}

%%
%% If your work has an appendix, this is the place to put it.

\end{document}